\documentclass[twoside,leqno,twocolumn]{article}

\usepackage[letterpaper]{geometry}

\usepackage{ltexpprt}

\usepackage{amsmath}
\usepackage{graphicx}
\usepackage{subfigure}
\usepackage{color}
\usepackage{xcolor}
\usepackage{amssymb}
\usepackage{enumitem}

\newcommand{\Probab}[1]{\mbox{}{\bf{Pr}}\left[#1\right]}

\newcommand{\ExpectBracket}[1]{\mbox{}\langle#1\rangle}

\newcommand {\michael}[1]{{\color{red}\sf{[michael: #1]}}}

\usepackage[normalem]{ulem}

\usepackage[utf8]{inputenc} % allow utf-8 input
\usepackage[T1]{fontenc}    % use 8-bit T1 fonts
\usepackage{hyperref}       % hyperlinks
\usepackage{url}            % simple URL typesetting
\usepackage{booktabs}       % professional-quality tables
\usepackage{amsfonts}       % blackboard math symbols
\usepackage{nicefrac}       % compact symbols for 1/2, etc.
\usepackage{microtype}      % microtypography

\begin{document}

\title{Heavy-Tailed Universality Predicts Trends in Test Accuracies for Very Large Pre-Trained Deep Neural Networks
%\thanks{Supported by GSF grants ABC123, DEF456, and GHI789.}
}

\author{%
  Charles H. Martin%
  \thanks{Calculation Consulting,
  8 Locksley Ave, 6B,
  San Francisco, CA 94122.
  \texttt{charles@CalculationConsulting.com} }
  \and
  Michael W. Mahoney%
  \thanks{ICSI and Department of Statistics,
  University of California at Berkeley,
  Berkeley, CA 94720,
  \texttt{mmahoney@stat.berkeley.edu} }
}

\date{}

\maketitle

\begin{abstract}
\noindent
Given two or more Deep Neural Networks (DNNs) with the same or similar architectures, and trained on the same dataset, but trained with different solvers, parameters, hyper-parameters, regularization, etc., can we predict which DNN will have the best test accuracy, and can we do so without peeking at the test data? 
In this paper, we show how to use a new Theory of Heavy-Tailed Self-Regularization (HT-SR) to answer this. 
HT-SR suggests, among other things, that modern DNNs exhibit what we call Heavy-Tailed Mechanistic Universality (HT-MU), meaning
that the correlations in the layer weight matrices can be fit to a power law (PL) with exponents that lie in common Universality classes from Heavy-Tailed Random Matrix Theory (HT-RMT).
From this, we develop a Universal capacity control metric that is a weighted average of PL exponents. 
Rather than considering small toy NNs, we examine over 50 different, large-scale pre-trained DNNs, ranging over 15 different architectures, trained on ImagetNet, each of which has been reported to have different test accuracies. 
We show that this new capacity metric correlates very well with the reported test accuracies of these DNNs, looking across each architecture (VGG16/.../VGG19, ResNet10/.../ResNet152, etc.).
%Moreover, we show 
We also show 
how to approximate the metric by the more familiar Product Norm capacity measure, as the average of the log Frobenius norm of the layer weight matrices.
Our approach requires no changes to the underlying DNN or its loss function, it does not require us to train a model (although it could be used to monitor training), and it does not even require access to the ImageNet data.
\end{abstract}

%\vspace{-2mm}

\section{Introduction}
\label{sxn:intro}

%\vspace{-1mm}

We are interested in the following general question.
\vspace{-2mm}
\begin{itemize}[leftmargin=*]
\item
Given two or more Deep Neural Networks (DNNs) with the same or similar architectures, 
%%and 
trained on the same dataset, but trained with different solvers, parameters, hyper-parameters, regularization, etc., can we predict which DNN will have the best test accuracy, and can we do so without peeking at the test data? 
\end{itemize}
\vspace{-1mm}

\noindent
This question is both theoretical and practical. 
Theoretically, solving this would help to understand why this class of machine learning (ML) models performs as well as it does in certain classes of applications.
Practically, there are many motivating examples.
Here are two.
%%SPACE%% several.
\vspace{-2mm}
\begin{itemize}[leftmargin=*]
\item
\textbf{Automating architecture search.}
Developing DNN models 
%%often 
requires significant architecture engineering, so there is interest in automating the design of DNNs.
Current methods can produce a series of DNNs subject to given general architecture constraints, but the models must be evaluated using cross validation (CV).
DNNs have so many adjustable parameters that even when using CV it is possible to leak information from the test sets into the training data, thus producing brittle, non-robust models.
It is thus of interest to have design principles and quality metrics that do not depend on the test data and/or the~labels. 
\item
\textbf{Fine-Tuning Pre-trained Models.}
Since one often does not have enough labeled data to train a large DNN from scratch, many modern engineering solutions can re-use widely-available pre-trained DNNs, fine-tuning them on smaller data sets. 
This technique often works extremely well for visual tasks, using DNNs pre-trained on ImageNet; and recently it has become feasible for complex natural language processing (NLP) tasks. 
Sometimes, however, these fine-tuned models become brittle and non-robust---due to overtraining, because information leaks from the test set into the training data.
Here, it would also be very helpful to be able to fine-tune large, pre-trained DNNs without needing to peek at the test~data.
%%SPACE%% \item
%%SPACE%% \textbf{Unsupervised Deep Learning.}
%%SPACE%% \michael{Charles, put in sentences; or remove and ``several'' to ``two'' above.}
\end{itemize}
\vspace{-1mm}

To predict trends in the generalization accuracy of a series of DNN architectures, VC-like theories offer theoretical bounds on the generalization accuracy. 
Practically, such capacity metrics can guide the
theoretical development of new regularizers for traditional ML optimization problems (e.g., counterfactual expected risk minimization~\cite{JMLR:v16:swaminathan15a}), but the bounds themselves are far too loose to be used directly. 
Moreover, since the early days of NN research, 
it was known that 
%%even Vapnik himself argued that his 
VC theory could (probably) not be directly applied to the seemingly wildly non-convex optimization problem implicitly posed by NNs. 
(This has caused some researchers to suggest we need to rethink regularization in DNNs entirely.)

In light of this, 
%%Still perhaps it is not really this hard?  Recent work by 
Liao et al.~\cite{LMBx18_TR} used an appropriately-scaled, data-dependent Product Norm capacity control metric to bound the worst-case generalization error for several small (non production-quality, but still interesting) DNN models, and they showed that the bounds are remarkably tight.
There is, in fact, a large body of work on norm-based capacity control metrics, both recent, e.g.,~\cite{LMBx18_TR, SHNx17_TR,PLMx18_TR} and~\cite{NTS14_TR,NTS15,NBMS17_TR,BFT17_TR,YM17_TR,KKB17_TR,NBS17_TR,AGNZ18_TR,ACH18_TR,ZF18_TR}, as well as much older ~\cite{Bar97,MN09_TR}. 
Much of this work has been motivated by the observation that parameter counting and more traditional VC-based bounds tend to lead to 
vacuous results for modern state-of-the-art DNNs, e.g., since modern DNNs are heavily over-parameterized and depend so strongly on the training data.

As with most theoretical studies, Liao et al.'s approach and intent differ greatly from ours.
They seek \emph{worst-case} complexity bounds, motivated to reconcile discrepancies with more traditional statistical learning theory, and they apply them (to quite small-scale NNs).
To address our main question, we seek an \emph{average-case} or \emph{typical case} (for realistic large-scale NNs) complexity metric, viable in production to guide the development of better DNNs at scale.
%\michael{MM: do we want to call this typical case.}
Bounding a small toy model does not necessarily mean that the individual weight matrix norms in production-quality DNNs will be directly comparable.
In particular, it does not mean that we can directly compare the individual weight matrix norms across layers in different, and more complex, architectures. 
Also, 
%to achieve their result, 
Liao et al. had to modify the DNN optimization loss function.
This means that their approach cannot be tested/evaluated on any existing \emph{pre-trained} DNN architecture, e.g., the VGG and ResNet models, widely-used today in industry. 
Still, their results do \emph{suggest} that a Product Norm may work well as a \emph{practical} capacity metric for large, and perhaps even pre-trained, production-quality DNNs.   
We will evaluate this and show that it does.
More generally, to predict trends in the test accuracies, one needs some more \emph{Universal} empirical metric that transfers across DNN architectures.

Recent work by Martin and Mahoney~\cite{MM18_TR,MM19_HTSR_ICML} suggests a Universal empirical metric to characterize the amount of \emph{Implicit Self-Regularization} and, accordingly, the generalization capacity, for a wide range of pre-trained DNNs.%
\footnote{A short version of~\cite{MM18_TR} is available as~\cite{MM19_HTSR_ICML}. The long version contains many more results and a much more detailed exposition.  }
The metric (defined below) involves the power law (PL) exponents, $\alpha$, of individual layer weight matrices, $\mathbf{W}$, as determined by fitting the Empirical Spectral Density (ESD), $\rho(\lambda)$, to a PL distribution.
Looking in detail at a series of models, like AlexNet, VGG, ResNet, etc, they observe that the (linear) layer weight matrices almost always follow a PL distribution, and 
%%the 
fitted PL exponents nearly all lie within a universal range $\alpha\in[2,5]$. 
Analysis of a small model (MinAlexNet) demonstrates that smaller PL exponents $\alpha$ correspond to better generalization.
Subsequent work~\cite{MM18_unpub_work} demonstrated Heavy-Tailed (HT) behavior in nearly every pre-trained architecture studied, e.g., across nearly $7500$ layer weight matrices (and 2D feature maps), including DNNs pre-trained for computer vision tasks on ImageNet, and for several different NLP~tasks.

When one observes good empirical PL fits of 
%%the 
ESDs of 
%%the 
correlations of layer weight matrices, we say the DNN \emph{exhibits Heavy-Tailed behavior}.
Motivated by these empirical observations, and using the Universality properties of Heavy-Tailed Random Matrix Theory (HT-RMT), Martin and Mahoney developed a theory of Heavy-Tailed Self-Regularization (HT-SR) for DNNs~\cite{MM17_TR,MM18_TR,MM19_HTSR_ICML}.
We build on and extend that theory~here.

In Statistical Physics, Universality of PL exponents is very non-trivial, and it suggests the presence of a deeper, underlying, \emph{Universal mechanism} driving the system dynamics~\cite{SornetteBook,BouchaudPotters03}.
It is this \emph{Heavy Tailed Mechanistic Universality} (HT-MU), as we call it, that originally motivated our study.  
HT-MU applies to the analysis of complicated systems, including many physical systems, traditional NNs~\cite{EB01_BOOK,nishimori01}, and even models of the dynamics of actual spiking neurons.
Indeed, the dynamics of learning in DNNs 
%(and perhaps real neurons as well) 
seems to resemble a system near a phase transition, such as the phase boundary of spin glass, or a system displaying Self Organized Criticality (SOC), or a Jamming transition~\cite{GSdx18_TR,SGd18_TR}. 
Of course, we can not say which mechanism, if any, is at play. 
Instead, we use the machinery of  HT-RMT as a stand-in for a generative model of the weight matrices in DNNs, to catalog and model the HT behavior of DNNs.%
\footnote{Perhaps the most well-known Universality in RMT is associated with the Gaussian Universality class, where the sum of many random variables drawn from a wide range of distributions is ``approximately Gaussian,'' e.g., in the sense that the sum approaches a suitably-normalized Gaussian distribution.  As briefly reviewed in Appendix~\ref{sxn:theory-review}, HT Universality makes analogous (but, admittedly, more complicated) statements for random variables drawn from distributions in which the tails decay more slowly than those in the Gaussian Universality class~\cite{MM18_TR,MM19_HTSR_ICML}.}
This Universality \emph{suggests} that we look for a \emph{Universal Capacity Control Metric}%
\footnote{To be clear, this metric is Universal, not in the sense that it applies ``universally'' to every possible DNN, but in the Statistical Physics sense~\cite{SornetteBook,BouchaudPotters03} that it applies to matrices within/across HT ``Universality''~classes.}
to address our main question.

Our main results are the following.
\vspace{-2mm}
\begin{itemize}[leftmargin=*]
\item
We evaluate the Product Norm capacity control metric on a wide range of large-scale pre-trained production-level DNNs, including 
%%the 
VGG and ResNet series,
%% of models, 
demonstrating that it correlates well with 
%%the 
reported average test accuracies across many series of models.
\emph{While norm-based metrics have been applied to small models,
to our knowledge, evaluating this metric to predict trends in test accuracies of large-scale pre-trained models has never (until now) been~reported.}
\item
We introduce a new methodology to analyze the performance of large-scale pre-trained DNNs, using a phenomena observed in HT-SR Theory.
%% that we call Heavy-Tailed Mechanistic Universality (HT-MU).
We construct a Universal capacity control metric to predict average DNN test performance.
This metric is a weighted average of layer PL exponents, $\hat{\alpha}$, weighted by the log%
\footnote{Throughout, we use log base 10.}
of the Spectral norm (i.e., maximum eigenvalue $\lambda^{max}$) of layer correlation matrices: 
$$
\hat{\alpha}=\sum_{l\in L}\alpha_{l}\log\lambda_{l}^{max}  .
$$
\item
We apply our Universal capacity control metric $\hat{\alpha}$ to a wide range of large-scale pre-trained production-level DNNs, including the VGG and ResNet series of models, as well as many others.
This 
%%Universal 
metric correlates very well with the reported average test accuracies across many series of pre-trained DNNs.
%% such as the VGG, ResNet, and other series of arcitectures.
%(without the need for re-training or looking at test data, although one could use the metrics as a training diagnostic).
\item
We provide a derivation for a relation between our Universal capacity control metric $\hat{\alpha}$ and the well known Product Norm capacity control metric, i.e, in the form of the average log of the squared Frobenius norm:
$$
\langle\log\Vert\mathbf{W}\Vert^{2}_{F}\rangle\approx \frac{1}{N_L}\sum_{l\in L} \alpha_{l}\log\lambda_{l}^{max}  .
$$
We do not make precise the error in ``$\approx$'' but our derivation makes clear that we expect the approximation to be good for smaller $\alpha$ and less good for larger $\alpha$.
%% We also show that
%% % (in addition to providing worst-case bounds on generalization quality for rather small NNs)
%% such norm-based capacity control metrics 
%% %%also 
%% correlate well with the average test accuracy in large-scale production-level pre-trained DNNs.
\end{itemize}
\vspace{-1mm}

There is a tradeoff here: our $\hat{\alpha}$ metric has two parameters ($\alpha$ and $\lambda^{max}$), as opposed to the Product Norm capacity control metric, which has one ($\Vert\cdot\Vert^{2}_{F}$), and it is more expensive to compute, but it does perform better. %
%% \footnote{One might wonder whether the $\hat{\alpha}$ metric just a variation of more familiar norm-based metrics.  Although we lack the space to describe it, our $\hat{\alpha}$ metric works more generally on actual DNNs when other norm-based metrics inspired by worst-case bounds fail.  In particular, we can identify counter examples, most notably in compressed DNN models~\cite{CWZZ17_TR}.  For compressed models, we have observed that the average Frobenius Norm increases with decreasing test error, whereas the average $\alpha$ decreases, as expected.}
Informally, as opposed to looking only at the ``size'' or ``shape'' of a model, e.g., as with a norm-based metric, the parameters $\lambda_{l}^{max}$ and $\alpha_{l}$ take into consideration both the size and the shape of the model.

For both our Universal $\hat{\alpha}$ metric and the Product Norm metric, our empirical results are, to our knowledge, the first time such theoretical capacity metrics have been reported to predict (trends in) the test accuracy for \emph{pre-trained production-level} DNNs.
In particular, this illustrates the usefulness of these norm-based metrics beyond smaller models such as MNIST, CIFAR10, and CIFAR100. 
Our 
results, including for both our Universal metric and the Product Norm metric we consider, can be reproduced with the \texttt{WeightWatcher} package%
%~\cite{weightwatcher_package}; 
\footnote{\url{https://pypi.org/project/WeightWatcher/}};
and our
results suggest that our ``practical theory'' approach is fruitful more generally for engineering good algorithms for realistic large-scale DNNs.

A shorter conference version of this paper, without the appendices, has appeared as~\cite{MM20_SDM}.

\vspace{-2mm}

\section{Brief Overview of Heavy-Tailed Self-Regularization}
\label{sxn:theory-review_abridged}

\vspace{-1mm}

Here, we briefly review Martin and Mahoney's Theory of Heavy-Tailed Self-Regularization (HT-SR)~\cite{MM18_TR,MM19_HTSR_ICML}.
See Appendix~\ref{sxn:theory-review} for more details.

Write the Energy Landscape (or optimization function, parameterized by $\mathbf{W}_{l}$s and $\mathbf{b}_{l}$s) for a typical DNN with $L$ layers, with activation functions $h_{l}(\cdot)$, and with $N\times M$ weight matrices $\mathbf{W}_{l}$ and biases $\mathbf{b}_{l}$, as:
\begin{equation*}
%PRESQUISH% E_{DNN}=h_{L}(\mathbf{W}_{L}\times h_{L-1}(\mathbf{W}_{L-1}\times h_{L-2}(\cdots)+\mathbf{b}_{L-1})+\mathbf{b}_{L})  .
E_{DNN} \hspace{-1mm} = \hspace{-1mm} h_{L}(\mathbf{W}_{L}\cdot h_{L-1}(\mathbf{W}_{L-1}\cdot h_{L-2}(\cdots)+\mathbf{b}_{L-1})+\mathbf{b}_{L})  .
%\label{eqn:dnn_energy}
\end{equation*}
%WLOG,
Typically, this model would be trained on some labeled data $\{d_{i},y_{i}\}\in\mathcal{D}$, using Backprop, by minimizing the loss $\mathcal{L}$.
For simplicity, we do not indicate the structural details of the layers (e.g., Dense or not, Convolutions or not, Residual/Skip Connections, etc.). 
%Each layer is defined by, e.g., one or more layer 2D weight matrices $\mathbf{W}_{l}$, and/or the 2D feature maps $\mathbf{W}_{l,i}$ extracted from 2D Convolutional (Conv2D) layers.

In the HT-SR Theory, we analyze the eigenvalue spectrum (the ESD) of the associated correlation matrices~\cite{MM18_TR,MM19_HTSR_ICML}.
From this, we can characterize the amount and form of correlation (and therefore the implicit self-regularizartion) present in the DNN's weight matrices.
For each layer matrix $\mathbf{W}_{l}$, of size $N \times M$, construct the associated $M\times M$ (uncentered) correlation matrix $\mathbf{X}_{l}$. 
Dropping the $L$ and $l,i$ indices, we have
$
\mathbf{X} = \frac{1}{N}\mathbf{W}^{T}\mathbf{W}.
$
If we compute the eigenvalue spectrum of $\mathbf{X}$, i.e., $\lambda_i$ such that
$  % $$
\mathbf{X}\mathbf{v}_{i}=\lambda_{i}\mathbf{v}_{i} , 
$  % $$
then the ESD of eigenvalues, $\rho(\lambda)$, is just a histogram of the eigenvalues.
Using HT-SR Theory~\cite{MM18_TR,MM19_HTSR_ICML}, we can characterize the \emph{correlations} in a weight matrix by examining its ESD, $\rho(\lambda)$.
It can be well-fit to a power law (PL) distribution, given as
$
\rho(\lambda)\sim\lambda^{-\alpha}  ,
$
which is (at least) valid within a bounded range of eigenvalues $\lambda\in[\lambda^{min},\lambda^{max}]$.  

%%In Statistical Physics, Universality 
%%arises in systems with very strong correlations, at or near a critical point or phase transition. 
%%It is characterized by measuring experimentally certain ``observables'' that display HT behavior, with common---or Universal---PL exponents. 
%%More importantly, it indicates that a specific Universal mechanism drives the underlying physical process, e.g., Self Organized Criticality, directed percolation, etc.~\cite{SornetteBook,BouchaudPotters03}. 
%%For this reason, we refer to the Universality observed in HT-SR, i.e., in the ESDs of (pre-trtained) DNN weight matrices, as \emph{Heavy-Tailed Mechanistic Universality~(HT-MU)}.

When we observe HT behavior in $\mathbf{W}$, or rather its correlation matrix $\mathbf{X}$, we essentially use HT-RMT as a generative model. 
We say that we \emph{model} $\mathbf{W}$ \emph{as if} it is a random matrix, $\mathbf{W}^{rand}(\mu)$, drawn from a Universality class of HT-RMT (i.e., VHT, MHT, or WHT, as defined below). 
To characterize this HT-MU behavior, we use a HT variant of RMT and use HT random matrices to elucidate different Universality classes.
Let $\mathbf{W}(\mu)$ be an $N \times M$ random matrix with entries chosen i.i.d. from
$$
\Probab{ W_{i,j} } \sim \frac{W_{0}^{\mu}}{|W_{i,j}|^{1+\mu}}  ,
$$
where $W_{0}$ is the typical order of magnitude of $W_{i,j}$, and where $\mu>0$. 
There are at least 3 different Universality classes
of HT random matrices, defined by the range $\mu$ takes on:
\begin{itemize}
\item $0<\mu<2$: VHT: Universality class of Very Heavy-Tailed (or L\'evy) matrices;
\item $2<\mu<4$: MHT: Universality class of Moderately Heavy-Tailed (or Fat-Tailed) matrices;
\item $4<\mu$: WHT: Universality class of Weakly Heavy-Tailed matrices.
\end{itemize}

%%HT-RMT provides more than HT Universality classes.
%%It also provides simple relations between the empirical observables, e.g., the PL exponent $\alpha$ and the maximum eigenvalue $\lambda^{max}$ of each $\mathbf{W}$, with the parameter(s) $\mu$ of our generative theory, i.e, of~HT-RMT.   
%%As described in Appendix~\ref{sxn:theory-review}, \emph{due to Heavy Tailed Mechanistic Universality (HT-MU)}, we expect 
%%$$
%%\text{VHT\;\&\;MHT:}\;\;\;\lambda^{max}\sim N^{4/\mu-1}  
%%$$
%%(where, for simplicity, $Q=1$)  
%%to hold for matrices in these HT Universality classes (as evidenced by their ESD properties), e.g., DNN weight matrices $\mathbf{W}$ after training---\emph{even when the matrix is not itself a HT random matrix} and therefore not governed by RMT.
%%The $\alpha$ and $\lambda^{max}$ are empirically-measurable quantities---of real or synthetic matrices---while $\mu$ is a parameter of the HT-RMT model. 
%%We shall use these Universal HT finite-size relations to derive a simple capacity control metric for our HT-SR Theory, and relate this to the well known Product Norm capacity control metric.

\vspace{-2mm}

%\section{Using Heavy-Tailed Universality}
\section{Heavy-Tailed Mechanistic Universality and Capacity Control Metrics}
\label{sxn:theory-new}

\vspace{-1mm}

%\charlesX{\bf{Heavy-Tailed Self-Regularization and Capacity Control Metrics}}

%In this section, we will describe our proposed capacity control metric.
%
From prior work~\cite{MM18_TR,MM19_HTSR_ICML}, we expect that smaller PL exponents of the ESD imply more regularization and therefore better generalization. 
Since smaller norms of weight matrices often correspond to better capacity control~\cite{LMBx18_TR,SHNx17_TR,PLMx18_TR,BFT17_TR}, we would like to relate the empirical PL exponent $\alpha$ to the empirical Frobenius norm $\Vert\mathbf{W}\Vert_{F}$.
At least na\"{\i}vely, this is a challenge, since smaller PL exponents often correspond to larger matrix norms (and thus worse generalization!).
See Appendices~\ref{sxn:appendix-random-vs-real} and~\ref{sxn:appendix-universality} for more details.
To resolve this apparent discrepancy, we will exploit HT-MU to propose a Universal DNN complexity metric.

\paragraph{Form of a Proposed Universal DNN Complexity Metric.} 

The PL exponent $\alpha$ is a complexity metric for a single DNN weight matrix, with smaller values corresponding to greater regularization~\cite{MM18_TR,MM19_HTSR_ICML}.
% The fitted PL  ... AWK
It describes how well that matrix encodes complex correlations in the training data.
Thus, a natural class of complexity or capacity metrics to consider for a DNN is to take a \emph{weighted average}%
\footnote{There are several reasons we don't want an unweighted average: 
an unweighted average behaves differently for HT random matrices than for well-trained DNN weight matrices, and so it would not be Universal; 
we want a metric that relates the $\alpha$ of HT-SR Theory with known capacity control metrics such as norms of weight matrices, and including weights permits this flexibility; 
we want weights to encode information that ``larger'' matrices are somehow more important;
and unweighted averages, while sometimes providing predictive quality, do not perform as reliably~well. 
See Appendices~\ref{sxn:appendix-random-vs-real} and~\ref{sxn:appendix-universality} for more details.
}
of the PL exponents, $\alpha_{l,i}$, for each layer weight matrix $\mathbf{W}_{l,i}$:
\begin{equation}
\hat{\alpha}:=\dfrac{1}{N_L}\sum_{l,i}b_{l,i}\alpha_{l,i}  .
\label{eqn:alpha_hat_generic}
\end{equation}
Here, the smaller $\hat{\alpha}$, the better we expect the DNN to represent training data, and (presumably) the better the DNN will generalize.  % to new data.
The main question is: what are good weights~$b_{l,i}$?

As we now show, we can extract the weighted average $\hat{\alpha}$ directly from the more familiar Product Norm, by exploiting both HT Universality, and its finite-size effects, arising
in DNN weight matrices.

%%%\paragraph{THEOREM:} \emph{The data dependent VC-like complexity of a Deep Neural Network can be expressed a weighted average the of power law exponents describing the empirical spectral density of layer weight matrices}
%%
%%%\charles{\paragraph{PROOF:...}}

\paragraph{Product Norm Measures of Complexity.} 

%% XXX. I PUT THIS MOSTLY IN THE INTRO.
%% \charlesX{NEED TO CLARIFY 
%% Worst case Bounds vs Average Case for complexity metrics, and REVIEW MORE OF HIDary's work, either here and/or in the Intro}
%% \michael{Their method works well on toy data for worst-case, and to get it to work they need to modify the loss function in worse-case, but if we consider average case then we can apply it to large realistic DNNs---put these comments here or in intro.}

It has been suggested that the complexity, $\mathcal{C}$, of a DNN can be characterized by the product of the norms of layer weight matrices,
$$
\mathcal{C}\sim\Vert\mathbf{W}_{1}\Vert\times\Vert\mathbf{W}_{2}\Vert\cdots\Vert\mathbf{W}_{L}\Vert ,
$$
where $\Vert\mathbf{W}\Vert$ is, e.g., the Frobenius norm~\cite{LMBx18_TR, SHNx17_TR,PLMx18_TR}.
(Here, we can use either $\Vert\mathbf{W}\Vert$ or $\Vert\mathbf{W}\Vert^{2}$, and one can view $\mathcal{C}$ as akin to a data-dependent VC complexity.)
To that end, we consider a log~complexity
\begin{eqnarray*}
\log\mathcal{C} &\sim& \log\bigg[\Vert\mathbf{W}_{1}\Vert\times\Vert\mathbf{W}_{2}\Vert\cdots\Vert\mathbf{W}_{L}\Vert\bigg]  \\
                &\sim& \bigg[\log\Vert\mathbf{W}_{1}\Vert+\log\Vert\mathbf{W}_{2}\Vert\cdots\log\Vert\mathbf{W}_{L}\Vert\bigg]  ,
\end{eqnarray*}
and we define the average log norm of weight matrices (where $N_{L}$ is the number of layers)~as
\begin{equation}
\langle\log\Vert\mathbf{W}\Vert\rangle=\dfrac{1}{N_{L}}\sum_{l}\log\Vert\mathbf{W}_{l}\Vert  .
\label{eqn:av_log_norm}
\end{equation}

%% \michael{Ques: is the notation for layers or convolutions or what, be consistent with Eqn.~(\ref{eqn:alpha_hat_generic}).}
%% \charlesX{Need more references to Hidary's work}
%% MM: A BIT HERE, BUT MOST IS IN INTRO AND CONCLUSION.

\paragraph{A Universal, Linear, PL--Norm Relation.} 

Based on our empirical results and theoretical considerations, we propose a simple linear relation between the (squared) Frobenius norm $\Vert\mathbf{W}\Vert^{2}_{F}$ of $\mathbf{W}$, the PL exponent $\alpha$, and the maximum eigenvalue $\lambda^{max}$ of $\mathbf{X}$ (i.e., the spectral norm $\Vert\mathbf{X}\Vert_{2}=\frac{1}{N}\Vert\mathbf{W}\Vert^{2}_{2}$):  
\begin{equation}
\textbf{PL--Norm Relation:} 
\hspace{1mm} %\quad 
\alpha\log\lambda^{max}\approx\log\Vert\mathbf{W}\Vert^{2}_{F}  .
\label{eqn:basic_relation}
\end{equation}
To our knowledge, this is the first time this PL--Norm relation has been noted in the literature (although prior work has considered norm bounds for HT data~\cite{MN09_TR}).
A few comments on Eqn.~(\ref{eqn:basic_relation}).
First, it provides a connection between the PL parameter $\alpha$ of HT-SR Theory and the weight norm $\Vert\mathbf{W}\Vert^{2}_{F}$ of more traditional statistical learning theory.
Second, it has a structural form like that of the well-known Hausdorff dimension~\cite{Sch07}.
Third, it shows that PL exponents can alternatively be interpreted (up to the $\frac{1}{N}$ scaling) as the Stable Rank in Log-Units:
$$
\mbox{Log-Units Stable Rank:} 
\quad
\mathcal{R}^{log}_{s}:=\dfrac{\log\Vert\mathbf{W}\Vert^{2}_{F}}{\log\lambda^{max}}  \approx \alpha  .
$$
Our justification for proposing Eqn.~(\ref{eqn:basic_relation}) is three-fold.
%%\charles{GOOD}
\begin{enumerate}
\item
\label{enum:first}
We derive Eqn.~(\ref{eqn:basic_relation}) in the special case of very small PL exponent, $\alpha \rightarrow 1$ ($\mu\rightarrow 0$), for an $N\times M$ 
%random 
matrix $\mathbf{W}^{rand}(\mu)$ (with $N=M$, or $Q=1$, where $Q=N/M$).%
\footnote{In particular, while this is a limiting statement, we expect to observe small deviations from this when we are not in the limit.}
%%See Appendix~\ref{sxn:appendix-derivation-pl-norm-relation}.
\item
\label{enum:second}
For finite-size random matrices $\mathbf{W}^{rand}(\mu)$, we expect the MHT Universality class, $\mu\in(2,4)$, to behave \emph{like} the VHT Universality class, $\mu\in(1,2)$.
Because of this similarity, we expect that we can extend Eqn.~(\ref{eqn:basic_relation}), approximately, to larger PL exponents.
For $N\sim\mathcal{O}(100-1000)$, $\alpha\log\lambda^{max}$ increases nearly linearly with $\log\Vert\mathbf{W}^{rand}(\mu)\Vert^{2}_{F}$ as $\mu$ increases.
For larger $N$, the relation saturates for large $\mu$. 
See Appendix~\ref{sxn:appendix-finite-size}.
\item
\label{enum:third}
\emph{As evidence of HT-MU}, we observe empirically that Eqn.~(\ref{eqn:basic_relation}) also applies, approximately, to the real DNN weight matrices $\mathbf{W}$. 
We see that $\alpha\log\lambda^{max}$ is positively correlated with $\log\Vert\mathbf{W}\Vert^{2}_{F}$ as $\alpha$ increases, and even shows similar saturation effects at large $\alpha$.
See Appendix~\ref{sxn:appendix-random-vs-real}.
\end{enumerate}

%%SPACE MOVED TO APPENDIX.%% We will discuss each of these in more detail below.

Finally, based on Eqn.~(\ref{eqn:basic_relation}), we choose the weights in Eqn.~(\ref{eqn:alpha_hat_generic}) to be the log of the corresponding maximum eigenvalues of $\mathbf{X}$.
That is, for a given $l,i$, we have the weights in Eqn.~(\ref{eqn:alpha_hat_generic})~as
$$
b_{l,i} = \lambda_{l,i}^{max}  .
$$
Then, we define the complexity metrics for Linear and Convolutional Layers as follows:
%% $$
%% \text{Linear Layer:}\;\;\log\Vert\mathbf{W}_{L}\Vert^{2}_{F}\rightarrow\log\lambda^{max}_{L}\alpha_{L}  .
%% $$
%% \michael{Need to be consistent with superscripts and subscripts, on $\lambda$, in this par and elsewhere.}
%% $$
%% \text{Conv2D Layer:}\;\;\log\Vert\mathbf{W}_{L}\Vert^{2}_{F}\rightarrow \sum_{i=1}^{n_{L}}\log\lambda^{max}_{i,L}\alpha_{L,i}  .
%% $$
%% I CHANGED THIS EVERyWHERE %% \nred{Hard to read max..make superscript ?}
\begin{eqnarray*}
\text{Linear Layer:} & & \log\Vert\mathbf{W}_{l}\Vert^{2}_{F} 
%\quad 
\rightarrow 
%\quad 
\alpha_{l}\log\lambda_{l}^{max}  \\
\text{Conv2D Layer:} & & \log\Vert\mathbf{W}_{l}\Vert^{2}_{F} 
%\quad 
\rightarrow 
%\quad 
\sum_{i=1}^{n_{l}}\alpha_{l,i} \log\lambda_{l,i}^{max} , 
\end{eqnarray*}
where, for Conv2D Layers, we relate the ``norm'' of the 4-index Tensor $\mathbf{W}_{l}$ to the sum of the $n_{l}=c\times d$ terms for each feature map.
%% So, in the expression for the Product Norm for $\log\mathcal{C}$, we can replace each $\log\Vert\mathbf{W}_{L}\Vert$ term for layer $L$ with these above expressions, and take the average over all $N_{\alpha}$  matrices.  
This lets us compare the Product Norm to the weighted average of PL exponents as follows:
\begin{equation}
2\log\mathcal{C}
\hspace{-0.5mm}
=
\hspace{-0.5mm}
\langle\log\Vert\mathbf{W}\Vert^{2}_{F}\rangle 
%\quad 
\hspace{-0.5mm}
\rightarrow 
\hspace{-0.5mm}
%\quad 
\hat{\alpha} 
\hspace{-0.5mm}
:= 
\hspace{-0.5mm}
\dfrac{1}{N_{L}}
\hspace{-0.5mm}
\sum_{i,l}
\hspace{-0.5mm}
\alpha_{i,l}
\hspace{-0.5mm}
\log \lambda_{l,i}^{max}  .
\label{eqn:alpha_hat_specific}
\end{equation}
%% \michael{We are using subscripts in a slightly confusing way.}
%%
%% 
%% This expression resembles the more familiar Product Norm, but it accounts for finite-size effects that the Product Norm relation over-estimates.
%% \michael{Ques: is that true.} \charlesX{probably not}
%% 
%% We will see that our approach improves on the loose bound provided by the Product Norm, giving a more accurate expression for predicting trends in the average case test accuracy for real-world production-quality DNNs.
%%
Given these connections, in Section~\ref{sxn:emp}, we will use $\hat{\alpha}$ to analyze numerous pre-trained DNNs.

%% \charlesX{THE NEXT 2 PARAGRAPHICAL SECTIONS EXPAND ON THE ABOVE POINTS. THE FIRST DISCUSSES THE FACT THAT A RANDOM HT MATRIX HAS A DIVERGING NORM, WHEREAS THE CORRELATED MATRICES HAVE SMALLER NORMS.  THIS IS EXPECTED FROM THEORY ALSO (CITE THE CHICAGO GUYS).  

\paragraph{The PL--Norm Relation: Deriving a Special Case of Eqn.~(\ref{eqn:basic_relation}).}
%%\section{The PL--Norm Relation: Deriving a Special Case of Eqn.~(\ref{eqn:basic_relation})}
%%\label{sxn:appendix-derivation-pl-norm-relation}

Here, we derive Eqn.~(\ref{eqn:basic_relation}) in the special case of very small PL exponent, 
as 
%%$\alpha \rightarrow 1$, 
$\mu \rightarrow 0$, 
for an $N \times M$ random matrix $\mathbf{W}$, {with $M=N, Q=1$, and with elements drawn from Eqn.~(\ref{eqn:ht_dstbn}).%
\footnote{We derive Eqn.~(\ref{eqn:basic_relation}) at what is sometimes pejoratively known as ``at a physics level of rigor.''  That is fine, as our justification ultimately lies in our empirical results.  Recall our goal: to derive a very simple expression relating fitted PL exponents and Frobenius norms that is usable by practical engineers working with state-of-the-art models, i.e., not simply small toy models.  There is very little ``rigorous'' work on HT-RMT, less still on understanding finite-sized effects of HT Universality.  Hopefully, our results will lead to more work along these lines.  }
We seek a relation good in the region $\mu\in[0,2]$, and we will extend the $\mu\sim 0$ results to this full region.
That is, we establish this as an asymptotic relation for the VHT Universality class for very small~exponents.

To start, recall that 
%% $$ 
%% \Vert \mathbf{W}\Vert_{F}^{2}=\mbox{Trace}[\mathbf{W}^{T}\mathbf{W}]=N\;\mbox{Trace}[\mathbf{X}]  .
%% $$
$ 
\Vert \mathbf{W}\Vert_{F}^{2}=\mbox{Trace}[\mathbf{W}^{T}\mathbf{W}]=N\;\mbox{Trace}[\mathbf{X}]  .
$
Since, $\mu \gtrsim 0$, 
the eigenvalue spectrum is dominated by a single large eigenvalue, it follows that
$$
\Vert \mathbf{W}\Vert_{F}^{2}\approx N\lambda^{max}  , 
$$
where $\lambda^{max}$ is the largest eigenvalue of the matrix $\mathbf{X}$ (with the $1/N$ normalization).
Taking the log of both sides of this expression and expanding leads to
%\begin{eqnarray*}
%\log\Vert \mathbf{W}\Vert_{F}^{2} 
%   &\approx& \log \left( N\lambda^{max} \right) \\
%   &=&       \log N+\log\lambda^{max}  .
%\end{eqnarray*}
\begin{eqnarray*}
\log\Vert \mathbf{W}\Vert_{F}^{2} 
   \approx \log \left( N\lambda^{max} \right) 
   =       \log N+\log\lambda^{max}  .
\end{eqnarray*}
Rearranging, we get that 
$$
\dfrac{\log\Vert \mathbf{W}\Vert_{F}^{2}}{\log\lambda^{max}}\approx \dfrac{\log N}{\log\lambda^{max}}+1  .
$$
Thus, for a parameter $\alpha$ satisfying Eqn.~(\ref{eqn:basic_relation}), we have %%that 
$$
\alpha\approx \dfrac{\log N}{\log\lambda^{max}}+1  .
$$
%%%
%%%$$
%%%\alpha-1\approx \dfrac{\log N}{\log\lambda^{max}}  .
%%%$$
%%%
The relation between $\alpha$ and $\mu$ for the VHT Universality class is given in Eqn.~(\ref{eqn:alpha_mu_vht}) as
$ %$$
\alpha=\frac{1}{2}\mu+1  .
$ %$$
Thus, to establish our result, we need to show that
$$
\dfrac{\log N}{\log\lambda^{max}}\approx\dfrac{1}{2}\mu  .
$$
To do this, we use the relation of Eqn.~(\ref{eqn:scaling_of_lambda_max}) for the tail statistic, i.e., that 
$ %$$
\lambda^{max}\approx N^{4/\mu-1}  .
$ %$$
Taking the log of both sides gives
$$
\log\lambda^{max}\approx\log N^{4/\mu-1}=(4/\mu-1)\log N  ,
$$
from which it follows that
$$
\dfrac{\log N}{\log\lambda^{max}}\approx\dfrac{\log N}{(4/\mu-1)\log N}=\dfrac{1}{4/\mu-1}   .
$$
Finally, we can form the Taylor Series for $\dfrac{1}{4/\mu-1}$ around, e.g., $\mu=1.15\approx 1$, which gives 
$$
\dfrac{1}{4/\mu-1}\bigg\rvert_{\mu=1.15}\approx\dfrac{1}{2}\mu-\dfrac{1}{6}+\cdots\approx\dfrac{1}{2}\mu  .
$$
%% TAYLOR SERIES FIG %% This relation is depicted in Figure~\ref{fig:taylor-series}.
This establishes the approximate---and rather surprising---linear relation we want for $\mu\in[0,2]$ for
the VHT Universality class of HT-RMT.

%% TAYLOR SERIES FIG %% \begin{figure}[!htb]
%% TAYLOR SERIES FIG %%    \centering
%% TAYLOR SERIES FIG %%    \includegraphics[scale=0.18]{img/taylor-series.png} 
%% TAYLOR SERIES FIG %%   \caption{%
%% TAYLOR SERIES FIG %%            Taylor series expansion for $\frac{1}{4/\mu-1}$ at $\mu=1.15\approx1$.
%% TAYLOR SERIES FIG %%           }
%% TAYLOR SERIES FIG %%   \label{fig:taylor-series}
%% TAYLOR SERIES FIG %% \end{figure}

\vspace{-2mm}

\section{Empirical Results on Pre-trained DNNs}
\label{sxn:emp}

\vspace{-1mm}

Here, we summarize our empirical results.
%% for the VGG and ResNet series of models.
%%See Appendix~\ref{sxn:appendix-addl-empirical} for additional empirical results on other pre-trained DNN models.
%%
We only consider Linear and Conv2D layers because we 
%%will 
only examine series of commonly available, open source, pre-trained DNNs with these kinds of layers. 
All models have been trained on ImageNet, and reported test accuracies are widely available. 
Throughout, we use 
%%the 
Test Accuracies for the Top1 errors (where Accuracy = 100 - Top1 error).
We see similar results for 
%%the 
Top5 errors.
We emphasize that, \emph{for our analysis, we do not need to retrain these models---and we do not even need the test data!}

\paragraph{VGG and VGG\_BN Models.}

\begin{figure}[t] %[!htb]
   \centering
   \subfigure[log Frobenius norm $\langle\log\Vert\mathbf{W}\Vert_{F}\rangle$]{
      \includegraphics[scale=0.15]{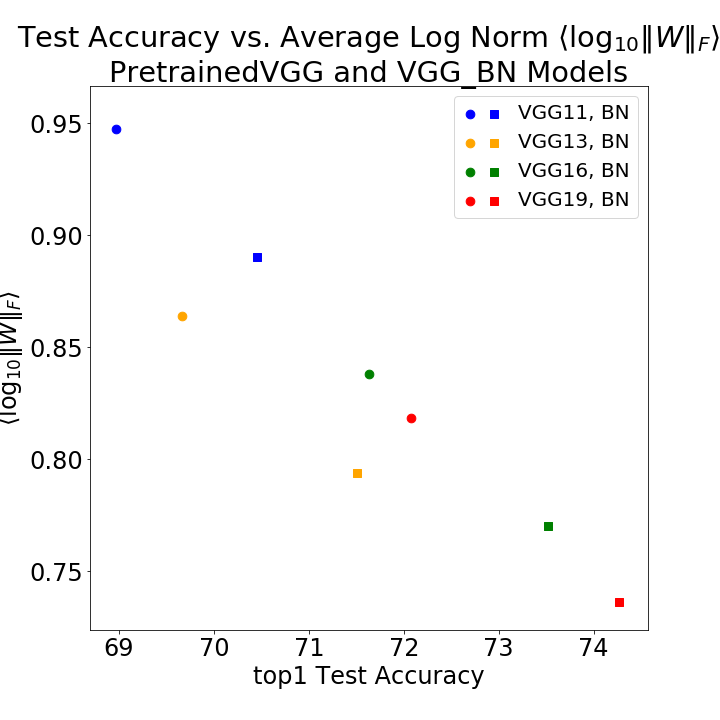}
      \label{fig:vgg_lognorms}
   }
   \subfigure[weighted average PL exponent $\hat{\alpha}$]{
      \includegraphics[scale=0.15]{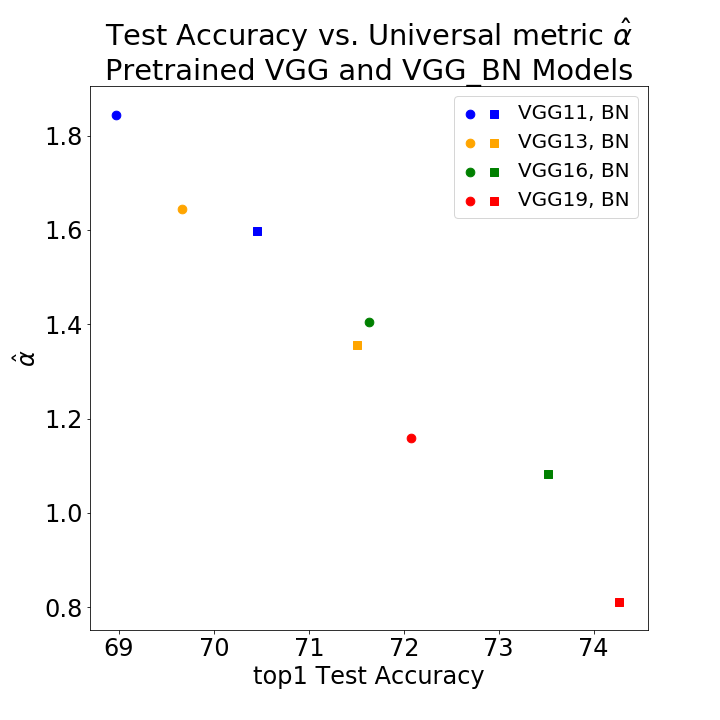}
      \label{fig:vgg_alphahat}
   }
   \caption{%
      Pre-trained VGG and VGG\_BN Architectures and DNNs.  
      Top 1 Test Accuracy versus
      average log Frobenius norm $\langle\log\Vert\mathbf{W}\Vert_{F}\rangle$ (in (\ref{fig:vgg_lognorms}))
      or
      Universal, weighted average PL exponent $\hat{\alpha}$ (in (\ref{fig:vgg_alphahat}))
      for
      VGG11 vs VGG11\_BN ({\color{blue}{blue}}),
      VGG13 vs VGG13\_BN ({\color{orange}{orange}}),
      VGG16 vs VGG16\_BN ({\color{green}{green}}),  and
      VGG19 vs VGG19\_BN ({\color{red}{red}}). 
      We plot plain the VGG models with circles and the VGG\_BN models with~squares.
   }
   \label{fig:vgg}
\end{figure}

%% COMBINED WITH ABOVE %% \begin{figure}[!htb]
%% COMBINED WITH ABOVE %%  \centering
%% COMBINED WITH ABOVE %%    \includegraphics[scale=0.40]{img/vgg-w_alphas.png}
%% COMBINED WITH ABOVE %%    \caption{
%% COMBINED WITH ABOVE %% Pre-trained VGG and VGG BN Architectures and DNNs.  Test Accuracy and weighted average $\hat{\alpha}$ for
%% COMBINED WITH ABOVE %%  VGG11 vs VGG11\_BN ({\color{blue}{blue}}),
%% COMBINED WITH ABOVE %% VGG13 vs VGG13\_BN ({\color{orange}{orange}}),
%% COMBINED WITH ABOVE %% VGG16 vs VGG16\_BN ({\color{green}{green}}),  and
%% COMBINED WITH ABOVE %% VGG19 vs VGG19\_BN ({\color{red}{red}}). 
%% COMBINED WITH ABOVE %% }
%% COMBINED WITH ABOVE %%   \label{fig:vgg_alphahat}
%% COMBINED WITH ABOVE %% \end{figure}

\begin{table}[t]
\small
\begin{center}
\begin{tabular}{|p{0.75in}|c|c|c|c|c|c|c|}
\hline
Model & Top1 Accuracy & $\hat{\alpha}$ \\
\hline
VGG11 & 68.97 & 1.84 \\
VGG11\_BN & 70.45 & 1.60 \\
\hline
VGG13 & 69.66 & 1.65 \\
VGG13\_BN & 71.51 & 1.36 \\
\hline
VGG16 & 71.64 & 1.41 \\
VGG16\_BN & 73.52 & 1.08 \\
\hline
VGG19 & 72.08 & 1.16 \\
VGG19\_BN & 74.27 & 0.81 \\
\hline
\end{tabular}
\end{center}
\caption{%
         Results for VGG Architecture.   Top1 Accuracy is defined
as the $100.0$ minus the Top1 reported error.
         }
\label{table:models_VGG}
\end{table}

We first look at the VGG class of models, comparing the log norm and the Universal $\hat{\alpha}$ metrics.
See Figure~\ref{fig:vgg} and Table~\ref{table:models_VGG} for a summary of the results.
Figures~\ref{fig:vgg_lognorms} and~\ref{fig:vgg_alphahat} show both the average log Frobenius norm, $\langle\log\Vert\mathbf{W}\Vert_{F}\rangle$ of Eqn.~(\ref{eqn:av_log_norm}), and the weighted average PL exponent, $\hat{\alpha}$ of Eqn.~(\ref{eqn:alpha_hat_specific}), as a function of the reported (Top1) test accuracy for the series of pre-trained VGG models, as available in the pyTorch package.%
\footnote{\url{https://pytorch.org/}}
These models include VGG11, VGG13, VGG16, and VGG19, as well as their more accurate counterparts with Batch Normalization, VGG11\_BN, VGG13\_BN, VGG16\_BN and VGG19\_BN. 
%See Figures~\ref{fig:vgg_lognorms} and~\ref{fig:vgg_alphahat} as well as Table~\ref{table:models_VGG} for details.
Table~\ref{table:models_VGG} provides additional details.

%Figure~\ref{fig:vgg_lognorms} shows the average log Frobenius norm results, which are quite good; and 
%Figure \ref{fig:vgg_alphahat} shows the weighted average PL exponent results, which yield slight improvements due to the method we introduce.
Across the entire series of architectures, 
reported test accuracies increase linearly as each metric, 
%%the average log Frobenius norm 
$\langle\log\Vert\mathbf{W}\Vert_{F}\rangle$
%%, Eqn.~(\ref{eqn:av_log_norm}), 
and 
%%the average weighted power law exponent 
$\hat{\alpha}$,
%%, Eqn.~(\ref{eqn}).
decreases.
Moreover, whereas the log norm relation has 2 outliers, VGG13 and VGG13\_BN, the Universal $\hat{\alpha}$ metric shows a near perfect linear relation across the entire VGG~series.

\paragraph{ResNet Models.}

\begin{figure}[!htb]
   \centering
   \subfigure[log Frobenius norm $\langle\log\Vert\mathbf{W}\Vert_{F}\rangle$]{
      \includegraphics[scale=0.12]{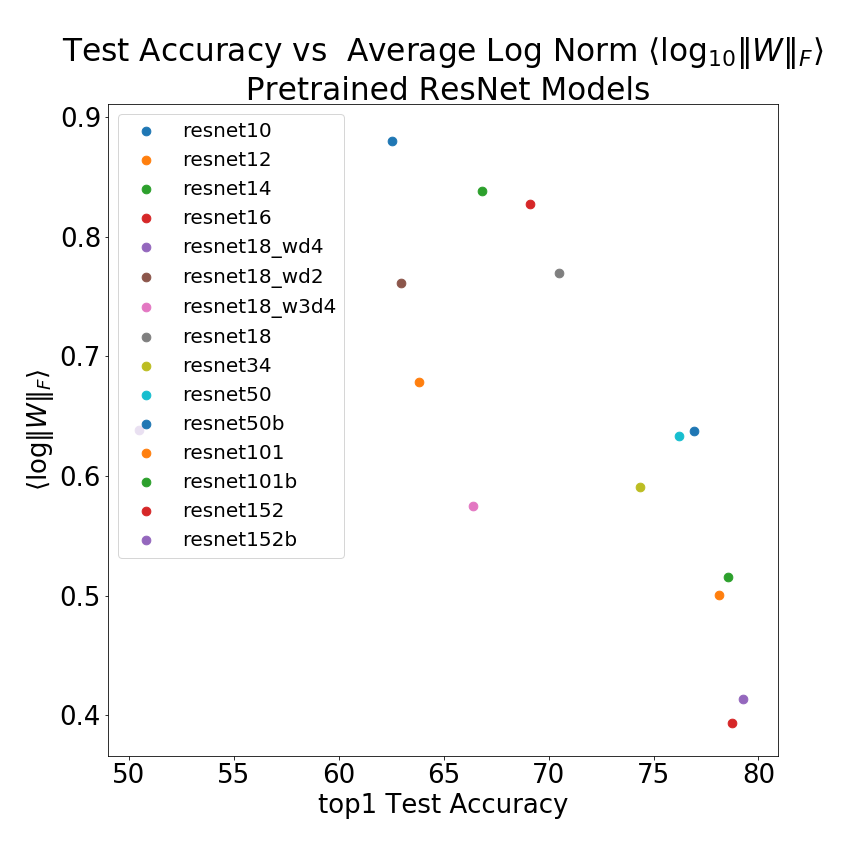}
      \label{fig:resnet_lognorms}
   }
   \subfigure[weighted average PL exponent $\hat{\alpha}$]{
      \includegraphics[scale=0.12]{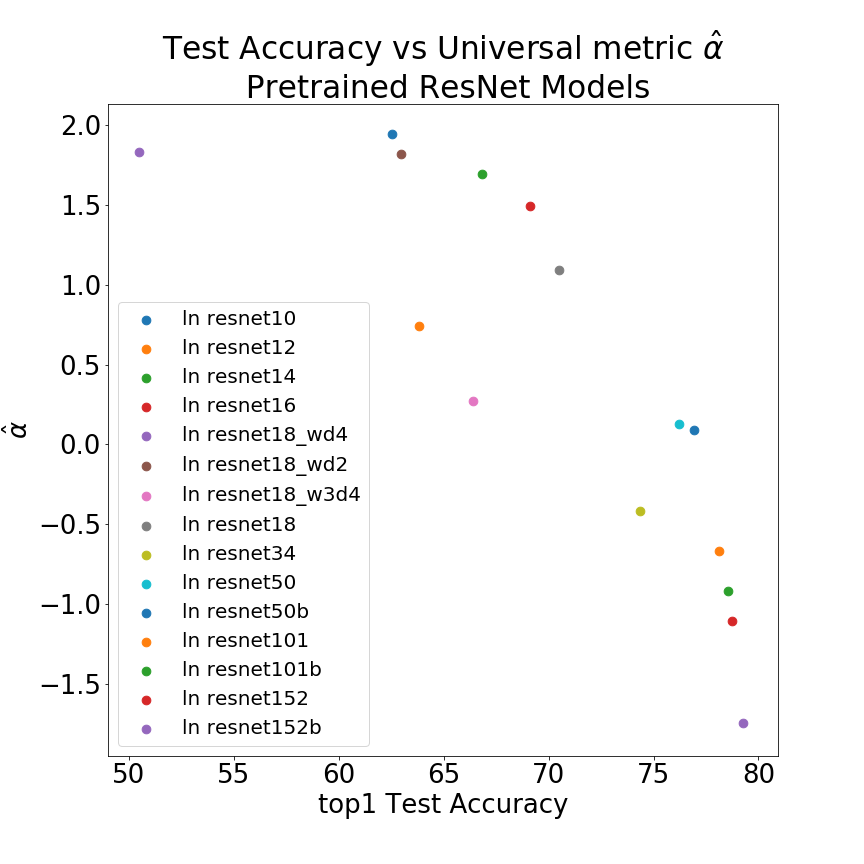}
      \label{fig:resnet_alphahat}
   }
   \caption{
      Pre-trained
      ResNet Architectures and DNNs.  
      Top 1 Test Accuracy versus
      average log Frobenius norm $\langle\log\Vert\mathbf{W}\Vert_{F}\rangle$ (in (\ref{fig:resnet_lognorms}))
      or
      Universal, weighted average PL exponent $\hat{\alpha}$ (in (\ref{fig:resnet_alphahat})).
           }
   \label{fig:resnet}
\end{figure}

\begin{table}[t] %[!htb]
\small
\begin{center}
\tabcolsep=0.11cm
\begin{tabular}{|p{0.75in}|c|c|c|c|c|c|c|}
\hline
Architecture 
 & Model
 & Top1 Accuracy & $\hat{\alpha}$ \\
\hline
ResNet (small)   & resnet10 & 62.54 & 1.94 \\
 & resnet12 & 63.82 & 0.74 \\
 & resnet14 & 66.83 & 1.70 \\
 & resnet16 & 69.10 & 1.49 \\
 \hline
 ResNet18  & resnet18\_wd4 & 50.50 & 1.83 \\
 & resnet18\_wd2 & 62.96 & 1.82 \\
 & resnet18\_w3d4 & 66.39 & 0.28 \\
 & resnet18 & 70.48 & 1.09 \\
 \hline
ResNet34 & resnet34 & 74.34 & -0.42 \\
\hline
ResNet50  & resnet50 & 76.21 & 0.13 \\
 & resnet50b & 76.95 & 0.09 \\
 \hline
ResNet101 & resnet101 & 78.10 & -0.67 \\
 & resnet101b & 78.55 & -0.92 \\
\hline
ResNet152 & resnet152 & 78.74 & -1.11 \\
 & resnet152b & 79.26 & -1.74 \\
\hline
\end{tabular}
\end{center}
\caption{Results for ResNet Architectures and DNN Models.  The Top1 Accuracy is defined
as the $100.0$ minus the Top1 reported error.  Some $\hat{\alpha}<0$ because the of how
the ResNet weight matrices are internally scale and normalized, which makes the maximum eigenvalue
less then one, $\lambda^{max}<1$.
        }
\label{table:models_resnet}
\end{table}

%\vspace{-1mm}

We next look at the ResNet class of models. 
See Figure~\ref{fig:resnet} and Table~\ref{table:models_resnet} for a summary of the results.
Here, we consider a set of 15 different pre-trained ResNet models, of varying sizes and accuracies, ranging from the small ResNet10 up to the largest ResNet152 models, as provided by the OSMR sandbox,%
\footnote{\url{https://github.com/osmr/imgclsmob}}
developed for training large-scale image classification networks for embedded systems.
Again, we compare the reported (Top1) test accuracy versus the average log norm $\langle\log\Vert\mathbf{W}\Vert_{F}\rangle$ and the Universal $\hat{\alpha}$ metrics. 

As with the VGG series, both metrics monotonically decrease as 
%%the 
test accuracies decrease for 
%%the 
ResNet series, and both metrics have a few large outliers off the main line relation. 
See Figures~\ref{fig:resnet_lognorms} and~\ref{fig:resnet_alphahat}.
In particular, the log norm metric has several notable outliers, including resnet18\_wd2, resnet18\_wd3\_d4, resnet34, and resnet10. 
The $\hat{\alpha}$ metric shows a slightly better relation, with resnet18\_wd2 more in line, and the other 3 outliers a little less off the main line of correlation. 
The 
%%Universal 
$\hat{\alpha}$ metric is as good or slightly better than 
%%the 
average log norm metric for the Resnet series of models. 

We see similar results for our Universal PL capacity control metric $\hat{\alpha}$ across a wide range of other pre-trained DNN models, described in next.  %%Appendix~\ref{sxn:appendix-addl-empirical}.
In nearly all cases, the metric $\hat{\alpha}$ correlates well with the reported test accuracies, with only a three DNN architectures as exceptions. 
Overall the $\hat{\alpha}$ metric systematically correlates well with the generalization accuracy of a wide class of pre-trained DNN architectures---which is rather remarkable.

%%\section{Additional Empirical Results}
%%\label{sxn:appendix-addl-empirical}
%%
%%In addition to the VGG and ResNet series of models, we examined a wide range of other DNNs.
%%Here, we summarize some of those results.

\paragraph{More Pre-trained Models.}

We present results for eleven more series of pre-trained DNN architectures, eight of which show positive results, as with the VGG and ResNet series,
%% (in Section~\ref{sxn:emp}), 
and three of which provide counterexample architectures.
See Table~\ref{table:models_more} for a summary.  % of~results.

The results that perform as expected are show in
Figures~\ref{fig:models_more_0}, 
\ref{fig:models_more_1}, 
\ref{fig:models_more_2}, 
and~\ref{fig:models_more_3}.
For each set of models, our Universal metric $\hat{\alpha}$
is smaller when, for the most part, the reported (Top 1)
test accuracy is larger. 
This holds approximately
true for the three of the four DenseNet models, with
densenet169 as an outlier. 
In fact, this is the only
outlier out of 26 DNN models in these 8 architectures.
For all of the other pre-trained DNNs, smaller
$\hat{\alpha}$ corresponds with smaller test error
and larger test accuracy, as predicted by our theory.

\begin{table}[!htb]
\small
\begin{center}
\tabcolsep=0.11cm
\begin{tabular}{|p{1in}|c|c|c|c|c|c|c|}
\hline
Architecture 
 & Model
 & Top 1 & $\hat{\alpha} $\\
 \hline
 Working & & & \\
 \hspace{2mm}Examples & & & \\
 \hline
 DenseNet
& densenet121 & 74.43 & 1.25 \\
& densenet161 & 77.14 & 0.84 \\
& densenet169 & 75.60 & 0.68 \\
& densenet201 & 76.90 & 0.50 \\
\hline
SqueezeNet
& squeezenet\_v1\_0 & 58.69 & 2.55 \\
& squeezenet\_v1\_1 & 58.18 & 1.56 \\
\hline
CondenseNet
& condensenet74\_c4\_g4 & 73.75 & -1.83 \\
& condensenet74\_c8\_g8 & 71.07 & -1.63 \\
\hline
DPN
& dpn68 & 75.83 & 0.57 \\
& dpn98 & 79.19 & 0.11 \\
& dpn131 & 79.46 & -0.13 \\
\hline
ShuffleNet
& shufflenetv2\_wd2 & 58.52 & 5.12 \\
& shufflenetv2\_w1 & 65.61 & 2.86 \\
\hline
MobileNet
& mobilenet\_wd4 & 53.74 & 5.54 \\
& mobilenet\_wd2 & 63.70 & 4.26 \\
& mobilenet\_w3d4 & 66.46 & 4.41 \\
& mobilenet\_w1 & 70.14 & 4.19 \\
& mobilenetv2\_wd4 & 50.28 & 12.12 \\
& mobilenetv2\_wd2 & 63.46 & 4.69 \\
& mobilenetv2\_w3d4 & 68.11 & 4.21 \\
& mobilenetv2\_w1 & 70.69 & 3.50 \\
\hline
SE-ResNet
& seresnet50 & 77.53 & -0.35 \\
& seresnet101 & 78.12 & -1.24 \\
& seresnet152 & 78.52 & -1.53 \\
\hline
SE-ResNeXt
& seresnext50\_32x4d & 79.00 & 1.81 \\
& seresnext101\_32x4d & 80.04 & 0.76 \\
\hline
Counter- & & &  \\
\hspace{2mm}examples & & &  \\
\hline
ResNeXt
& resnext101\_32x4d & 78.19 & 1.22 \\
& resnext101\_64x4d & 78.96 & 1.34 \\
\hline
MeNet
& menet108\_8x1\_g3 & 56.08 & 5.31 \\
& menet128\_8x1\_g4 & 56.05 & 4.46 \\
& menet228\_12x1\_g3 & 66.43 & 4.82 \\
& menet256\_12x1\_g4 & 66.59 & 4.97 \\
& menet348\_12x1\_g3 & 69.90 & 5.74 \\
& menet352\_12x1\_g8 & 66.69 & 4.42 \\
& menet456\_24x1\_g3 & 71.60 & 5.11 \\
\hline
FDMobileNet
& fdmobilenet\_wd4 & 44.23 & 6.40 \\
& fdmobilenet\_wd2 & 56.15 & 7.01 \\
& fdmobilenet\_w1 & 65.30 & 7.10 \\
\hline
\end{tabular}
\end{center}
\caption{Results for more pre-trained DNN models.  Models provided in the OSMR Sandbox, implemented in pyTorch. Top 1 refers to the Top 1 Accuracy, which $100.0$ minus the Top 1 reported error.}
\label{table:models_more}
\end{table}

\begin{figure}[!htb]
   \centering
   \subfigure[DenseNet] {
       \includegraphics[scale=0.19]{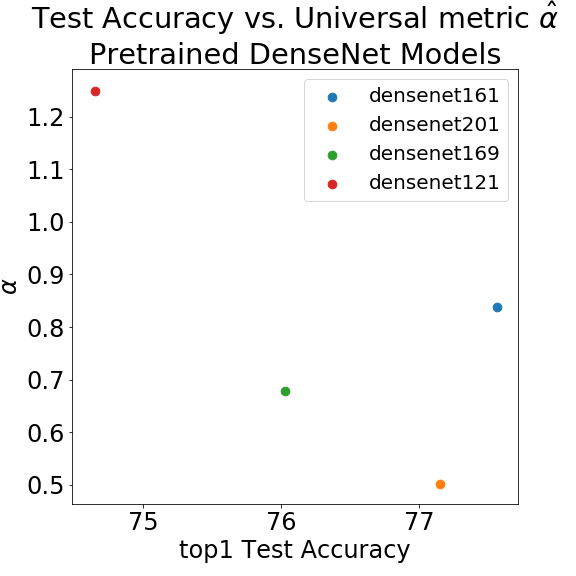} 
       \label{fig:densenet}
   }
   \subfigure[SqueezeNet]{
       \includegraphics[scale=0.19]{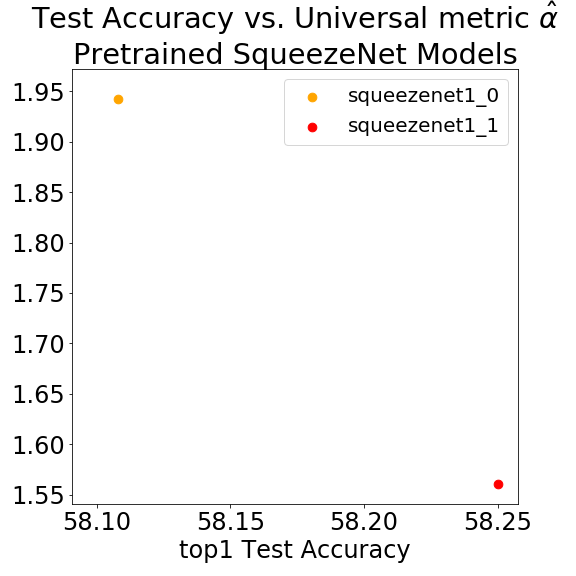} 
       \label{fig:squeezenet}
   }
   \caption{
      Pre-trained 
      Densenet and SqueezeNet PyTorch 
      Models.
      Top 1 Test Accuracy versus~$\hat{\alpha}$.
           }
   \label{fig:models_more_0}
\end{figure}

\begin{figure}[!htb]
   \centering
   \subfigure[CondenseNet]{
       \includegraphics[scale=0.15]{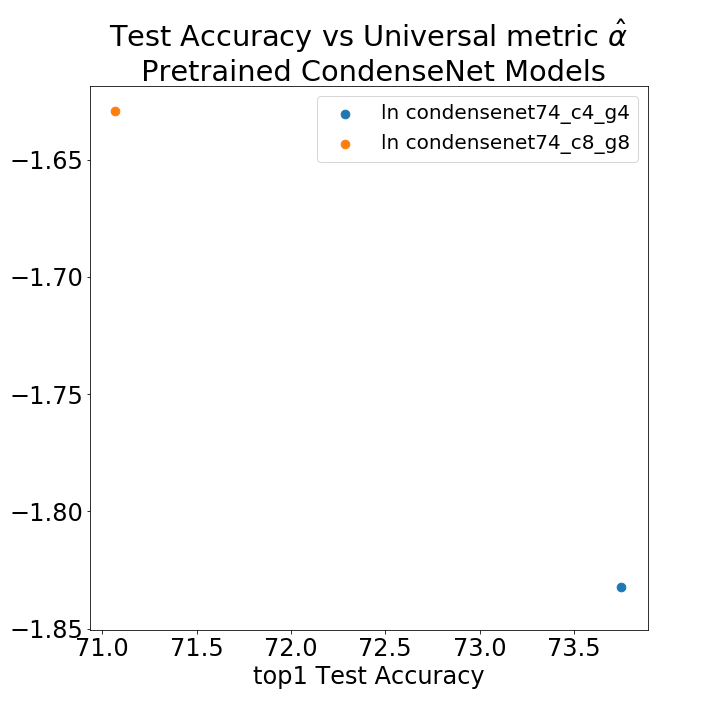} 
       \label{fig:densenet-small}
   }
   \subfigure[DPN]{
       \includegraphics[scale=0.15]{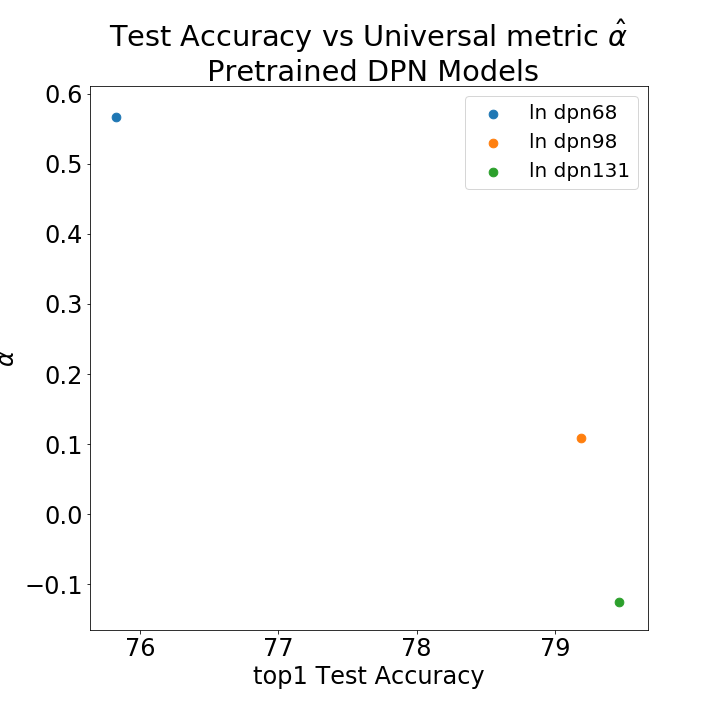} 
       \label{fig:dpn-net}
   }
   \caption{
      Pre-trained 
      CondenseNet and DPN
      Models.
      Top 1 Test Accuracy versus
      $\hat{\alpha}$.
           }
   \label{fig:models_more_1}
\end{figure}

\begin{figure}[!htb]
   \centering
   \subfigure[ShuffleNet]{
      \includegraphics[scale=0.15]{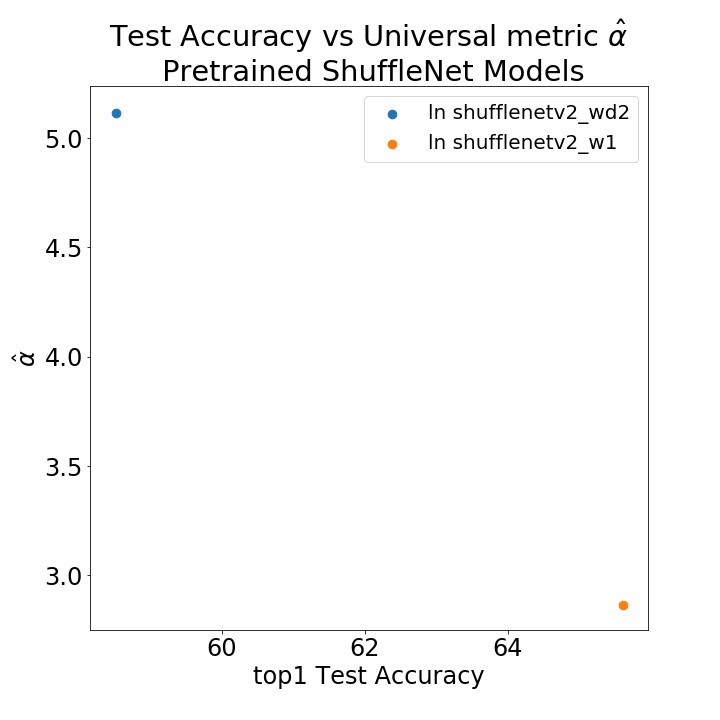}
      \label{fig:shufflenet-small_X1}
   }
   \subfigure[MobileNet]{
      \includegraphics[scale=0.15]{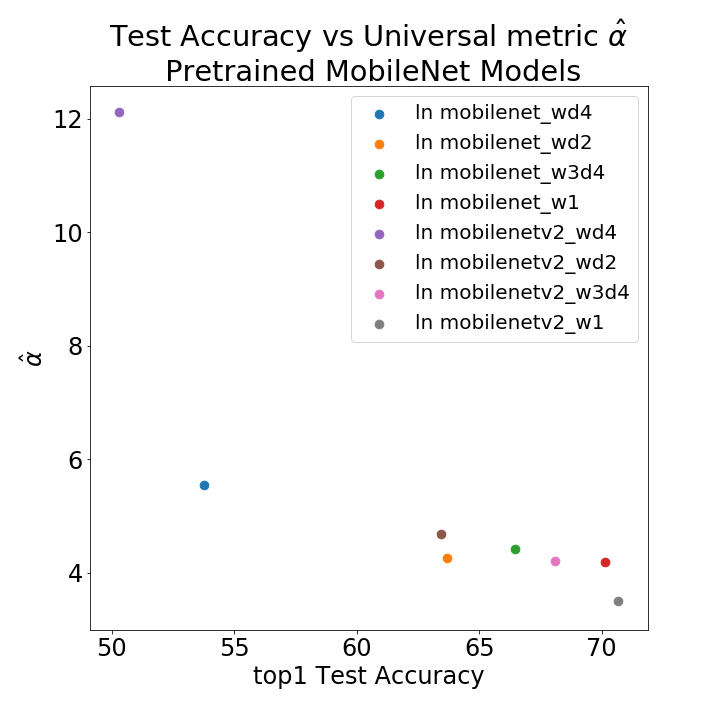} 
      \label{fig:resnet-small}
   }
   \caption{
      Pre-trained 
      ShuffleNet and MobileNet
      Models.
      Top 1 Test Accuracy versus
      $\hat{\alpha}$.
           }
   \label{fig:models_more_2}
\end{figure}

\begin{figure}[!htb]
   \centering
   \subfigure[SeResNet]{
      \includegraphics[scale=0.15]{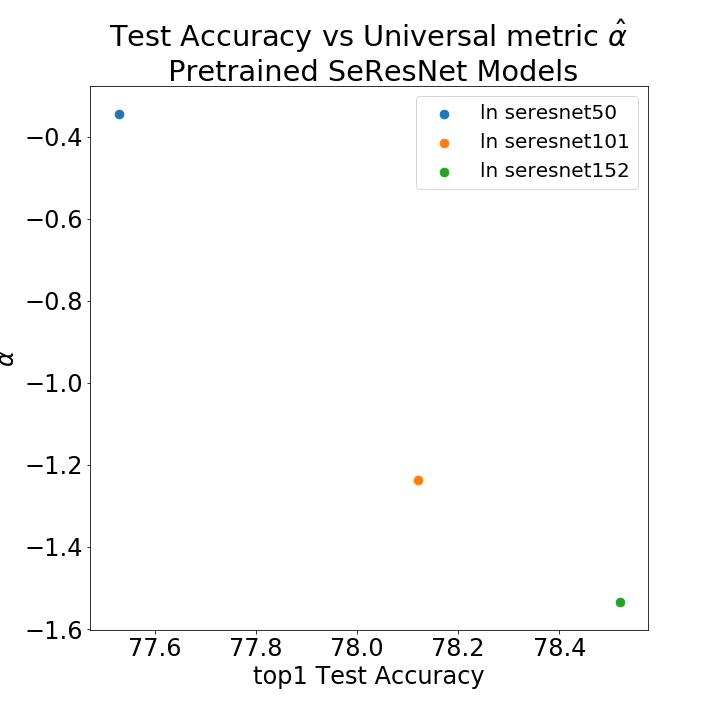}
      \label{fig:shufflenet-small_X2}
   }
   \subfigure[SeResNeXt]{
      \includegraphics[scale=0.15]{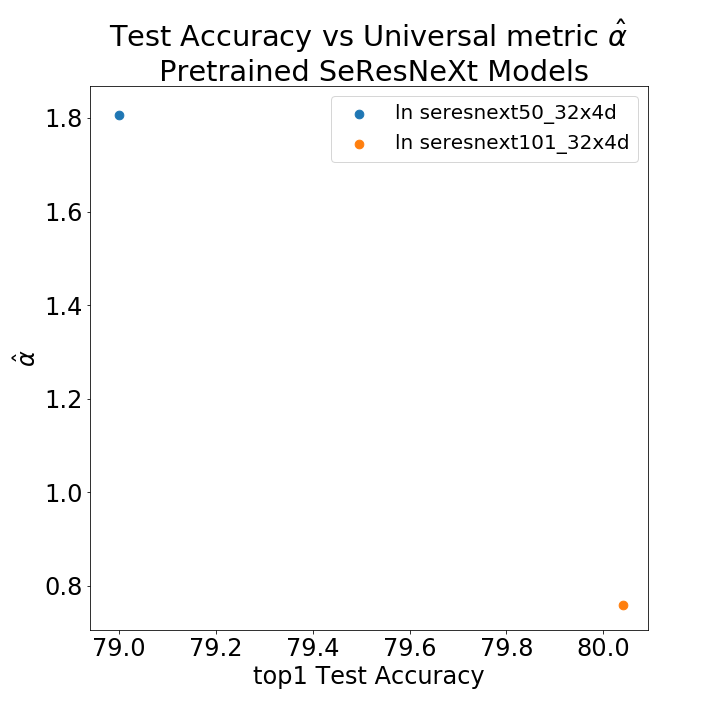}
      \label{fig:shufflenet-small_X3}
   }
   \caption{
      Pre-trained 
      SeResNet and SeResNeXt
      Models.
      Top 1 Test Accuracy versus
      $\hat{\alpha}$.
           }
   \label{fig:models_more_3}
\end{figure}

\paragraph{Counterexamples.}
In such a large corpus of DNNs, there are of course exceptions for a predictive theory.
See
%%Figure~\ref{fig:counter-examples}
%%(as well as the corresponding rows of 
Table~\ref{table:models_more}
%%)
for the counterexamples.
These are ResNeXt, MeNet, and FDMobileNet.
For ResNeXt, there are only two models, and the $\hat{\alpha}$ is larger
for the less accurate model. 
For MeNet, there are seven different models,
and there is no discernible pattern in the data. 
Finally, for FDMobileNet,
there are three different pre-trained models, and, again, the 
$\hat{\alpha}$ is larger for the less accurate models. 
We have not looked in detail at these results and simply present them for completeness. 
 
%%\begin{figure*}[!htb]
%%   \centering
%%   \subfigure[ResNeXt]{
%%      %\includegraphics[scale=0.19]{img/ResNeXt-w_alphas.png} 
%%      \includegraphics[scale=0.21]{img/ResNeXt-w_alphas.png} 
%%      \label{fig:resnet-small}
%%   }
%%   \subfigure[MeNet]{
%%      %\includegraphics[scale=0.19]{img/MeNet-w_alphas.png} 
%%      \includegraphics[scale=0.21]{img/MeNet-w_alphas.png} 
%%      \label{fig:menet-net}
%%   }
%%   \subfigure[FDMobileNet]{
%%      %\includegraphics[scale=0.19]{img/FDMobileNet-w_alphas.png} 
%%      \includegraphics[scale=0.21]{img/FDMobileNet-w_alphas.png} 
%%      \label{fig:resnet-small}
%%   }
%%   \caption{
%%      Pre-trained 
%%      ResNeXt, MeNet, and FDMobileNet
%%      Models 
%%      provide counterexamples to our main trends.
%%      Top 1 Test Accuracy versus
%%      $\hat{\alpha}$.
%%           }
%%   \label{fig:counter-examples}
%%\end{figure*}

\vspace{-2mm}
\section{Discussion and Conclusion}
\label{sxn:discussion}
\vspace{-1mm}

We have presented an \emph{unsupervised} capacity control metric which predicts trends in test accuracies of a trained DNN---without peeking at the test data. 
%%In the interests of space, 
See Appendix~\ref{sxn:appendix-addl-discussion} for more discussion.
%%We conclude by observing simply that 
%We expect our result will have applications in the fine-tuning of DNN hyperparameters as well as related challenges.
%Moreover, because we do not need to peek at the test data, our approach may prevent information from leaking from the test set into the model, thereby helping to prevent overtraining and making fined-tuned DNNs more robust.
%Finally, 
Our work 
%%also 
leads to a harder theoretical question: 
can one %%is it possible to 
characterize properties of realistic DNNs to determine whether a DNN is overtrained---without peeking at the test data?

\newpage
\bibliographystyle{unsrt}

{\small
%\bibliography{gen_gap}
%\bibliography{dnns}
\bibliography{dnns,gen_gap}
}

\appendix
\newpage

%\vspace{-2mm}

\section{Overview of Heavy-Tailed Self-Regularization}
\label{sxn:theory-review}

%\vspace{-1mm}

%In this section, we will briefly review results from Universality that will be relevant for our analylsis.
We review Martin and Mahoney's Theory of Heavy-Tailed Self-Regularization (HT-SR)~\cite{MM18_TR,MM19_HTSR_ICML}.
%%\michael{We need to push the other 10 pager to the archive, without the supplementary information}

Write the Energy Landscape (or optimization function) for a typical DNN with $L$ layers, with activation functions $h_{l}(\cdot)$, and with $N\times M$ weight matrices $\mathbf{W}_{l}$ and biases $\mathbf{b}_{l}$, as:
\begin{equation*}
%PRESQUISH% E_{DNN}=h_{L}(\mathbf{W}_{L}\times h_{L-1}(\mathbf{W}_{L-1}\times h_{L-2}(\cdots)+\mathbf{b}_{L-1})+\mathbf{b}_{L})  .
E_{DNN} \hspace{-1mm} = \hspace{-1mm} h_{L}(\mathbf{W}_{L}\cdot h_{L-1}(\mathbf{W}_{L-1}\cdot h_{L-2}(\cdots)+\mathbf{b}_{L-1})+\mathbf{b}_{L})  .
%\label{eqn:dnn_energy}
\end{equation*}
%WLOG,
Typically, this model would be trained on some labeled data $\{d_{i},y_{i}\}\in\mathcal{D}$, using Backprop, by minimizing the loss $\mathcal{L}$.
For simplicity, we do not indicate the structural details of the layers (e.g., Dense or not, Convolutions or not, Residual/Skip Connections, etc.). 
Each layer is defined by one or more layer 2D weight matrices $\mathbf{W}_{l}$, and/or the 2D feature maps $\mathbf{W}_{l,i}$ extracted from 2D Convolutional (Conv2D) layers.
(We have not yet analyzed LSTM or other complicated Layers.) 
A typical modern DNN may have anywhere between 5 and 5000 2D layer matrices.%
\footnote{%
%Some notational conventions:
For each Linear Layer, we get a  single $(N\times M)$ (real-valued) 2D weight matrix, denoted $\mathbf{W}_{l}$, for layer $l$.  
This includes Dense or Fully-Connected (FC) layers, as well as 1D Convolutional (Conv1D) layers, Attention matrices, etc.
We ignore the bias terms $\mathbf{b}_{l}$ in this analysis. 
%XXX.  WHY, WHAT.   
Let the aspect ratio be $Q=\frac{N}{M}$, with $Q\ge 1$.
For the Conv2D layers, we have a 4-index Tensor, of the form $(N\times M \times c\times d)$, consisting
of $c\times d$ 2D feature maps of shape $(N\times M)$.    
We  extract $n_{l}=c\times d$ 2D weight matrices $\mathbf{W}_{l,i}$, one for each feature map $i=[1,\dots,n_{l}]$ for layer $l$.
%One could construct matrices in other ways.
}

\paragraph{Heavy-Tailed Empirical Spectral Distributions.}
%
%By Universal behavior, we mean that the eigenvalue spectrum associated weight matrices 
%We can identify different Universality classes 
%
In the HT-SR Theory, we analyze the eigenvalue spectrum (the ESD) of the associated correlation matrices~\cite{MM18_TR,MM19_HTSR_ICML}.
From this, we can characterize the amount and form of correlation, and therefore implicit self-regularizartion, present in the DNN's weight matrices.
For each layer weight matrix $\mathbf{W}$, of size $N \times M$, construct the associated $M\times M$ (uncentered) correlation matrix $\mathbf{X}$. 
Dropping the $L$ and $l,i$ indices, we have
$$
\mathbf{X} = \frac{1}{N}\mathbf{W}^{T}\mathbf{W}.
$$
%In theoretical treatments, $\gamma$ depends on the form of $\Probab{W_{i,j}}$, e.g., the value of $\mu$, in order to prove the existence of the limiting forms of the ESD.
%For MP theory and the Gaussian Universality class, we can set $\gamma=1$, but for the HT Universality classes, we need to set $\gamma=2/\mu$.
%Of course, empirically, we do not know the PL exponent $\mu$, or the particular Universality class, \emph{a priori}, and the data are of only finite size.
%This will be very important below. 
%For our empirical analysis, we set $\gamma=1$ and deal with these issues in an \emph{a posteriori} manner. 
%
If we compute the eigenvalue spectrum of $\mathbf{X}$, i.e., $\lambda_i$ such that
$  % $$
\mathbf{X}\mathbf{v}_{i}=\lambda_{i}\mathbf{v}_{i} , 
$  % $$
then the ESD of eigenvalues, $\rho(\lambda)$, is just a histogram of the eigenvalues, formally written as
\begin{equation}
\rho(\lambda)=\sum\limits_{i=1}^{M}\delta(\lambda-\lambda_{i})  .
\label{eqn:eigenval_hist}
\end{equation}
%
%From HT-RMT theory~\cite{XXX-XXX,XXX-XXX,XXX-XXX,XXX-XXX}, the ESD $\rho(\lambda)$ of a HT matrix will have a HT, taking the form
%% POSSIBLY REDUNDANT %% Empirically, for weight matrices of a modern well-trained production-quality DNN, the ESD nearly always exhibits Heavy-Tailed properties~\cite{MM18_TR,MM19_HTSR_ICML}.
%%SPACE%% \footnote{Older and smaller models exhibit ESD properties closer to a Spiked Covariance model, and it is possible to train DNNs to have ESDs with other properties, corresponsing to other types of implicit self regularization~\cite{MM18_TR,MM19_HTSR_ICML}.} 
Using HT-SR Theory, we can characterize the \emph{correlations} in a weight matrix by examining its ESD, $\rho(\lambda)$.
It can be well-fit to a power law (PL) distribution, given~as
\begin{equation}
\rho(\lambda)\sim\lambda^{-\alpha}  ,
\label{eqn:eigenval_pl}
\end{equation}
which is (at least) valid within a bounded range of eigenvalues $\lambda\in[\lambda^{min},\lambda^{max}]$.  
%\michael{Ques: here, $\alpha$ is theoretical, while below $\alpha$ is fit, so clarify.}
%
We can determine $\alpha$ by fitting the   ESD to a PL, using the commonly accepted Maximum Likelihood (MLE) method of Clauset et al.~\cite{CSN09_powerlaw,ABP14}.
%(((
%\charles{Discuss fact the PL tails are Frechet at finite-size so only need to fit the bulk}
%\michael{Ques: clarify.}
%)))
This method works very well for exponents between $\alpha\in(2,4)$; and it is adequate, although imprecise, for smaller and especially larger $\alpha$~\cite{newman2005_zipf}. 

The original work on HT-SR Theory~\cite{MM18_TR,MM19_HTSR_ICML} considered NNs including AlexNet and InceptionV3 (as well as DenseNet, ResNet, and VGG), and it showed that for nearly every $\mathbf{W}$, the (bulk and tail) of the ESDs can be fit to a PL and the PL exponents $\alpha$ nearly all lie within the range $\alpha\in(1.5,5)$.
Moreover, smaller exponents $\alpha$ are correlated with more implicit self-regularization and, correspondingly, better generalization~\cite{MM18_TR,MM19_HTSR_ICML}.
Subsequent work~\cite{MM18_unpub_work} has shown that these results are ubiquitous.
%\michael{Put ref to notebook in~\cite{MM18_unpub_work}.}
For example, 
upon examining nearly 10,000 layer weight matrices $\mathbf{W}_{l,i}$ across over 50 different modern pre-trained DNN architectures, the ESD of nearly every $\mathbf{W}$ layer matrix can be fit to a PL:
$70-80\%$ of the time, the fitted PL exponent $\alpha$ lies in the range $\alpha\in(2,4)$; and  %% (in the Moderately Heavy-Tailed Universality class, described below); and
$10-20\%$ of the time, the fitted PL exponent $\alpha$ lies in the range $\alpha< 2$.  %% (in the Very Heavy-Tailed Universality class, described below).
For example, see Figure~\ref{fig:power-law-histogram} for a histogram of results for ca. 7500 weight matrices from ImageNet.
%\charlesX{explain this plot, ref to notebook and check in.  This plot is for ImageNet, includes conv@D feature maps, lots of small alpha, but not as good a fit.}
Of course, there are exceptions: in any real DNN, the fitted $\alpha$ may range anywhere from $\sim 1.5$ to $10$ or higher~\cite{MM18_unpub_work} (and, of course, larger values of $\alpha$ may indicate that the PL is not a good model for the data).  
Still, overall, in nearly all large, pre-trained DNNs, the correlations in the  weight matrices exhibit a remarkable Universality, being both Heavy Tailed, and having small---but not too small---PL exponents. 

\begin{figure}[!htb]
   \centering
   \includegraphics[scale=0.40]{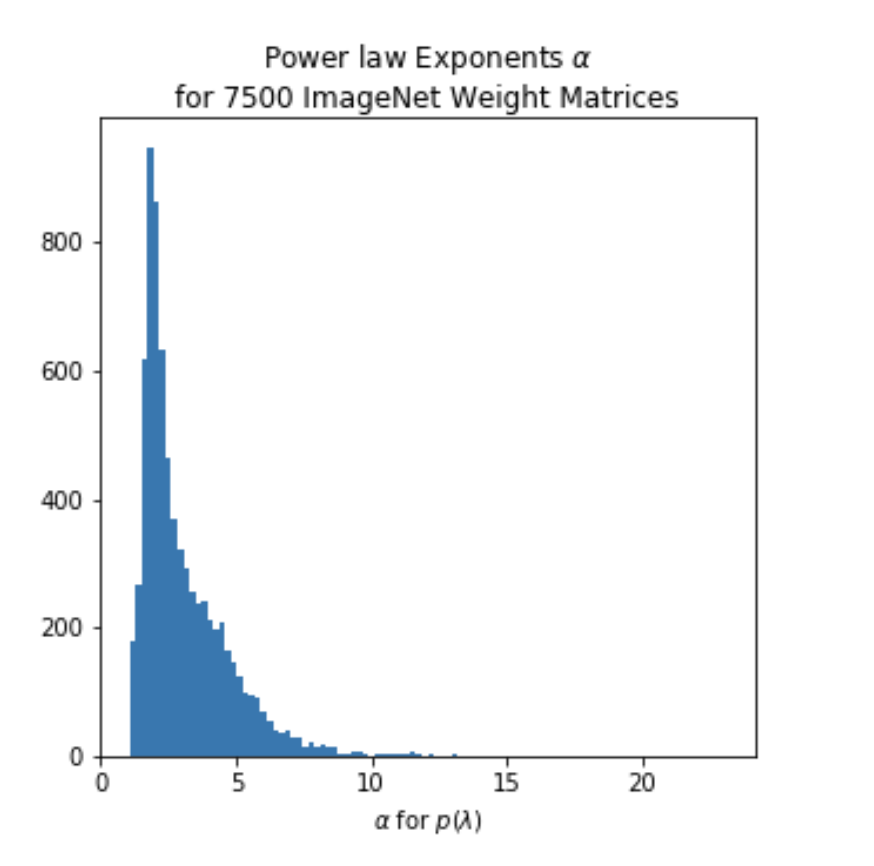} 
   \caption{%
            Histogram of PL exponents, $\alpha$, fit on ca. 7500 Linear and Conv2D Layers from ImageNet. 
            The vast majority have $\alpha\in(1.5,5)$. 
            Some Conv2D layers have smaller values of $\alpha$; and larger values of $\alpha$ (up to ca. 20) exist but correspond to less reliable fits.
            }
   \label{fig:power-law-histogram}
\end{figure}

%\nred{removed this: \paragraph{Simple Random Matrix Models.} }
%One might imagine that the matrix elements of $\mathbf{W}$ are drawn from some probability distribution, e.g., a Normal $N(0,\sigma)$ distribution
%\begin{equation}
%\Probab{ W_{i,j} } \sim N(0,\sigma)
%\end{equation}
%with mean $0$ and variance $\sigma$.%
%\footnote{At the start of training, DNN weight matrices typically are approximately Normal.  This has been used by some as an analytically-tractable model for trained DNNs, but it is an empirical question whether this is a good model for very non-random matrices that arise at the end of training modern DNNs.  Empirically, it is not~\cite{MM18_TR,MM19_HTSR_ICML}.}
%%MM%% \charlesX{Our imagination here lets us derive seemingly \emph{Universal} expressions for how the correlations should behave, even though $\mathbf{W}$  itself is not at all random. }
%These Gaussian models arise in the well known Marchenko-Pastur (MP) RMT~\cite{XXX-XXX}, as well as the Spiked-Covariance model~\cite{johnstone2009}, which is a perturbative variant of MP-RMT. 
%Empirically, it is know that these Gaussian-based models are useful for understanding older, smallish Neural Networks, such as LeNet5~\cite{MM18_TR,MM19_HTSR_ICML}.
%More modern DNNs, however, display very different, more exotic,  Universal, behavior.  

\paragraph{Heavy-Tailed Mechanistic Universality.} 

%\charlesX{THIS SECTION NEEDS A LOT OF WORK}

Here, we consider the question:
what does it \emph{mean} to say that DNN weight matrices exhibit Universality? 
This answer to this---and its implications---depends on your perspective, in particular, as a Mathematician or a Physicist.

In Statistics and Applied Mathematics, Universality typically refers to properties of systems that can be modeled by random matrices.
The justification is that certain system properties can be deduced, without requiring knowledge of system details, from a few global quantities that are then used as parameters to define a random matrix ensemble~\cite{ER05,EW13}.
Of course, the DNN weight matrices are not random matrices---they are strongly-correlated objects---so it may seem odd that we can apply RMT to characterize them.

In Statistical Physics, Universality refers to a different, but related, phenomena.
It arises in systems with very strong correlations, at or near a critical point or phase transition. 
It is characterized by measuring experimentally certain ``observables'' that display HT behavior, with common---or Universal---PL exponents. 
More importantly, it indicates that a specific Universal mechanism drives the underlying physical process, e.g., Self Organized Criticality, directed percolation, etc.~\cite{SornetteBook,BouchaudPotters03}. 
For this reason, we refer to the Universality observed in HT-SR, i.e., in the ESDs of (pre-trtained) DNN weight matrices, as \emph{Heavy-Tailed Mechanistic Universality~(HT-MU)}.

For an illustration of what we mean by HT-MU, see Figure~\ref{fig:universality_diagram}. 
When we observe HT behavior in $\mathbf{W}$, or rather its correlation matrix $\mathbf{X}$, we use HT-RMT as a generative model. 
We say that we \emph{model} $\mathbf{W}$ \emph{as if} it is a random matrix, $\mathbf{W}^{rand}(\mu)$, drawn from a Universality class of HT-RMT (i.e., VHT, MHT, or WHT, as defined below). 
Of course, we do not mean that $\mathbf{W}$ is itself random in any way.
We simply use RMT as a stand-in generative model because the correlations in $\mathbf{W}^{rand}(\mu)$ resembles the correlations in  $\mathbf{W}$. 
Indeed, RMT itself does not describe the behavior of a random matrix $\mathbf{W}$ per-se, but it characterizes its correlations (i.e., the eigenvalues of $\mathbf{X}$). 
Specifically, RMT describes the ESD, $\rho_{emp}(\lambda)$, its limiting, deterministic form $\rho_{\infty}(\lambda)$, as well as the finite-size scaling and fluctuations of the maximum eigenvalue, $\lambda^{max}$~\cite{MM18_TR,MM19_HTSR_ICML}. 
So, even though $\mathbf{W}$ is not random, we expect its ESD and maximum eigenvalue to behave \emph{as if} they were drawn from some HT-random matrix $\mathbf{W}^{rand}(\mu)$.

\begin{figure}[t]  %[!htb]
   \centering
   \includegraphics[scale=0.36]{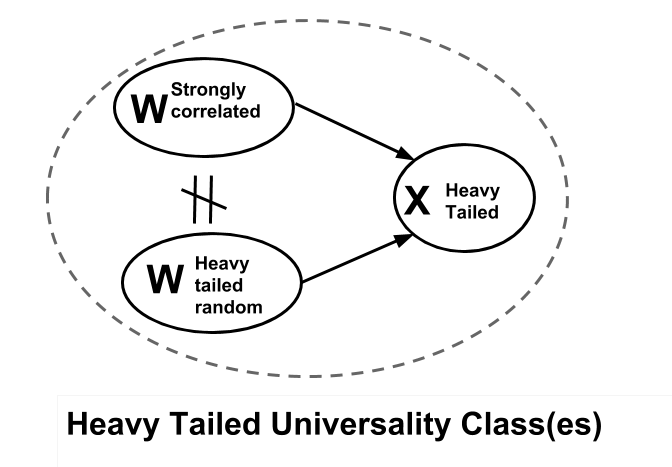} 
   \caption{Illustration of our use of HT Universality.  
            %\michael{Charles, maybe tweak this figure to rotate and horizontally show transfer of Universal things between different matrices.}  
            A random matrix $\mathbf{W}(\mu)$ with elements drawn i.i.d. from the HT distribution of Eqn.~(\ref{eqn:ht_dstbn}) and weight matrix $\mathbf{W}^{str. corr.}$ from a modern well-trained DNN that exhibits strong correlations each arise from different generative mechanisms, and they are different.  Both exhibit similar Universality properties, as evidenced by the HT properties of the ESDs of their corresponding correlation matrices $\mathbf{X}$, and so we expect Universal properties to be similar between~them.}
   \label{fig:universality_diagram}
\end{figure}

\paragraph{Heavy-Tailed Random Matrix Theory.} 
To characterize this HT-MU behavior, we use a HT variant of RMT and use HT random matrices to elucidate different Universality classes.
Let $\mathbf{W}(\mu)$ be an $N \times M$ random matrix with entries chosen i.i.d. from
\begin{equation}
\Probab{ W_{i,j} } \sim \frac{W_{0}^{\mu}}{|W_{i,j}|^{1+\mu}}  ,
\label{eqn:ht_dstbn}
\end{equation}
where $W_{0}$ is the typical order of magnitude of $W_{i,j}$, and where $\mu>0$. 
These HT matrix models were first introduced in the Statistical Physics literature, where they are called L\'evy Matrices when $0<\mu<2$~\cite{PB94}; see also~\cite{BM97,BJNx01_TR,BJNx06_TR,heavytails2007}.
More recently, there has been mathematical work on HT random matrices~\cite{AG08,AAP09,BJ09_TR,DPS14,AT16}.
There are at least 3 different Universality classes%
\footnote{Results for $\mu=2,4$ are slightly different~\cite{SornetteBook,BouchaudPotters03}.  We don't describe them since we don't expect to be able to resolve them numerically.  Also, sometimes L\'evy matrices are split into VHT for $1<\mu<2$ and EHT (Extremely Heavy-Tailed) for $0<\mu<1$, as the properties for these two parameter regimes are somewhat different~\cite{SornetteBook,BouchaudPotters03}.}
of HT random matrices, defined by the range $\mu$ takes on:
\begin{itemize}
\item $0<\mu<2$: VHT: Universality class of Very Heavy-Tailed (or L\'evy) matrices;
\item $2<\mu<4$: MHT: Universality class of Moderately Heavy-Tailed (or Fat-Tailed) matrices;
\item $4<\mu$: WHT: Universality class of Weakly Heavy-Tailed matrices.
\end{itemize}
%%
%%\charles{ THIS SECTION REVIEWS Heavy Tail RMT, and is ALL REVIEW.  WE MUST  explain how to interpret a correlated ESD in terms of HT-RMT.  BUT be clear that this is not the whole story }
%%
%%\michael{We need to highlight how the HT Universality we use is one that crosses Universality classes, i.e., VHT and finite-sized MHT, and certainly is NOT limited to random matrices, but we use RMT to characterize certain things.}
%%

\paragraph{Heavy-Tailed, Finite-Size Relations.}

HT-RMT provides more than HT Universality classes.
It also provides simple relations between the empirical observables, e.g., the PL exponent $\alpha$ and the maximum eigenvalue $\lambda^{max}$ of each $\mathbf{W}$, with the parameter(s) $\mu$ of our generative theory, i.e, of~HT-RMT.   

%% For the VHT Universality class, the PL tail of Eqn.~(\ref{eqn:eigenval_pl}) persists in the infinite limit $N\rightarrow\infty$, for $Q$ fixed; and
%% we have the linear relation between our observed exponent $\alpha$ and the theoretical~$\mu$:
%% \begin{equation}
%% \alpha=\frac{1}{2}\mu+1  .
%% \label{eqn:alpha_mu_vht}
%% \end{equation}
%% \michael{Ques: careful, two things change here: observed versus theoretical, and matrix elements versus eigenvalues.}
%% This expression which works very well at finite size, even for very small matrices $(M,N\approx100)$.
%% %
%% For the MHT Universality class, the PL tail of Eqn.~(\ref{eqn:eigenval_pl}) vanishes in the infinite limit $N\rightarrow\infty$, for $Q$ fixed.
%% \michael{Ques: what does ``vanishes'' mean, it changes slope, and it is MP in the limit for WHT.}
%% At all finite sizes, however, it persists, and it follows a Frechet distribution (i.e., an exponentially-truncated PL). 
%% Here, $\alpha$ is still linear in $\mu$, but it displays very strong finite-size effects, empirically giving 
%% \begin{equation}
%% \alpha=a\mu+b, 
%% \label{eqn:alpha_mu_mht}
%% \end{equation}
%% where $a,b$ depend strongly on $M,N$. 
%% (See Table 3 of \cite{MM18_TR} and Figure~\ref{XXX} below for more details on this.)
%% These strong finite-size effects characterize these MHT distributions; and they are well-known in the Statistical Physics literature~\cite{SornetteBook,BouchaudPotters03}. 
%% We will exploit these finite-size effects to develop our-theory.

For the VHT Universality class, the PL tail of Eqn.~(\ref{eqn:eigenval_pl}) persists in the infinite $N\rightarrow\infty$ limit, for $Q$ fixed; and
we have the linear relation between our observed exponent $\alpha$ and the theoretical~$\mu$:
\begin{subequations}
\label{eqn:alpha_mu_vht_and_mht}
\begin{align}
\text{VHT:}\;\;\;\alpha=\frac{1}{2}\mu+1  .
\label{eqn:alpha_mu_vht}
\end{align}
%% \michael{Ques: careful, two things change here: observed versus theoretical, and matrix elements versus eigenvalues.}
This asymptotic expression works very well at finite size, even for very small matrices $(M,N\approx100)$.

For the MHT Universality class, the PL tail of Eqn.~(\ref{eqn:eigenval_pl}) holds in the infinite $N\rightarrow\infty$ limit, for $Q$ fixed, for $\alpha$ in Eqn.~(\ref{eqn:alpha_mu_vht}).
At all finite sizes, however, $\alpha$ is still linear in $\mu$, but it displays \emph{very} strong finite-size effects, empirically~giving: 
\begin{align}
\text{MHT:}\;\;\;\alpha=a\mu+b, 
\label{eqn:alpha_mu_mht}
\end{align}
\end{subequations}
where $a,b$ depend strongly on $M,N$. 
(See Table 3 of \cite{MM18_TR} for more details.)
These strong finite-size effects characterize MHT distributions; and they are well-known in Statistical Physics~\cite{SornetteBook,BouchaudPotters03}. 
We will exploit these finite-size effects to develop our~theory.

Finally, for both the VHT and the MHT Universality classes, the maximum empirical eigenvalue, $\lambda^{max}$, 
follows a Frechet distribution (i.e., an exponentially-truncated PL); and we expect it to scale with $N$ according to Extreme Value Theory (EVT)~\cite{heavytails2007,disordered2007,Resnick07,MM18_TR,MM19_HTSR_ICML}:
\begin{equation}
\text{VHT\;\&\;MHT:}\;\;\;\lambda^{max}\sim N^{4/\mu-1}  
\label{eqn:scaling_of_lambda_max}
\end{equation}
(where, for simplicity, $Q=1$).  
%%\charlesX{Add back in Q}

%Let's clarify Eqns.~(\ref{eqn:alpha_mu_vht_and_mht}) and~(\ref{eqn:scaling_of_lambda_max}):  
Eqns.~(\ref{eqn:alpha_mu_vht_and_mht}) and~(\ref{eqn:scaling_of_lambda_max}) show that we have very simple relationships that apply to random matrices that lie within both 
the VHT and MHT Universality classes.
The $\alpha$ and $\lambda^{max}$ are empirically-measurable quantities---of real or synthetic matrices---while $\mu$ is a parameter of the HT-RMT model. 
For us, the question is: how shall we \emph{use} these relations?

\emph{Due to Heavy Tailed Mechanistic Universality (HT-MU)}, we expect Eqn.~(\ref{eqn:scaling_of_lambda_max}) to hold for matrices in these HT Universality classes (as evidenced by their ESD properties), e.g., DNN weight matrices $\mathbf{W}$ after training---\emph{even when the matrix is not itself a HT random matrix} and therefore not governed by RMT or EVT.
%%% (See Appendix~\ref{sxn:appendix-derivation-two-relations} for a derivation of Eqns.~(\ref{eqn:alpha_mu_vht_and_mht}) and~(\ref{eqn:scaling_of_lambda_max}).)
We shall use these Universal HT finite-size relations to derive a simple capacity control metric for our HT-SR Theory, and relate this to the well known Product Norm capacity control metric.

%We will use this to derive a relationship that holds between $\alpha$, $\lambda^{max}$, and perhaps other observables such as matrix norms, and then 
%appeal to \nred{HT-MU} to apply it to  \nred{the highly} non-random, pre-trained DNN weight matrices that also lie within these HT Universality classes.

%%Appendix~\ref{sxn:appendix-finite-size}
%%Appendix~\ref{sxn:appendix-random-vs-real}
%%Appendix~\ref{sxn:appendix-universality}

%\newpage

%\paragraph{The PL--Norm Relation: Finite-Size Effects.}
\section{The PL--Norm Relation: Finite-Size Effects}
\label{sxn:appendix-finite-size}

Here, we consider finite-size effects in Eqn.~(\ref{eqn:basic_relation}), both within and across HT Universality classes, i.e., for both VHT and MHT matrices.
See Figure~\ref{fig:randW}, which  displays $\frac{\log\Vert\mathbf{W}\Vert^{2}_{F}}{\log\lambda^{max}}$ as a function of the fitted PL exponent $\alpha$, with varying sizes $N$ (with aspect ratio $Q=1$).
Recall that $\alpha \approx \frac{1}{2}\mu+1$ for VHT random matrices (Eqn.~(\ref{eqn:alpha_mu_vht})), while $\alpha = a\mu+b$ for MHT random matrices (Eqn.~(\ref{eqn:alpha_mu_mht})), where $a,b$ strongly depend on $N$ and $M$.
Thus, $\mu\in(0,2)$ for VHT matrices corresponds to $\alpha\in(1,2)$, while $\alpha \approx(2,5)$ for MHT matrices.
%% $$
%% \dfrac{\log\Vert\mathbf{W}\Vert^{2}_{F}}{\log\lambda^{max}}\;\;vs.\;\;(\alpha)  .
%% $$

The numerical results in Figure~\ref{fig:randW} show that as $\alpha$ increases
when $\alpha<2$, there exists a near-linear relation; and
when $\alpha>2$, for $N,M$ large, the relation saturates, becoming constant, while for smaller $N,M$, there exists a near-linear relation, but with strong finite-size effects.
These numerical results demonstrate that $ \log\Vert\mathbf{W}\Vert^{2}_{F}\approx\alpha\log\lambda^{max} $ works very well for VHT random matrices, for $\alpha<2$, and that it works moderately well for MHT matrices and even some WHT matrices.
In particular, for MHT matrices, in the finite-size regime, when $N,M\sim\mathcal{O}(100-1000)$, which is typical for modern DNNs, the PL-Norm relations holds, \emph{on average}, quite well. 
This is precisely what we want in a practical engineering metric that is designed to describe average test~accuracy. 

\begin{figure}[!htb]
   \centering
   \includegraphics[scale=0.15]{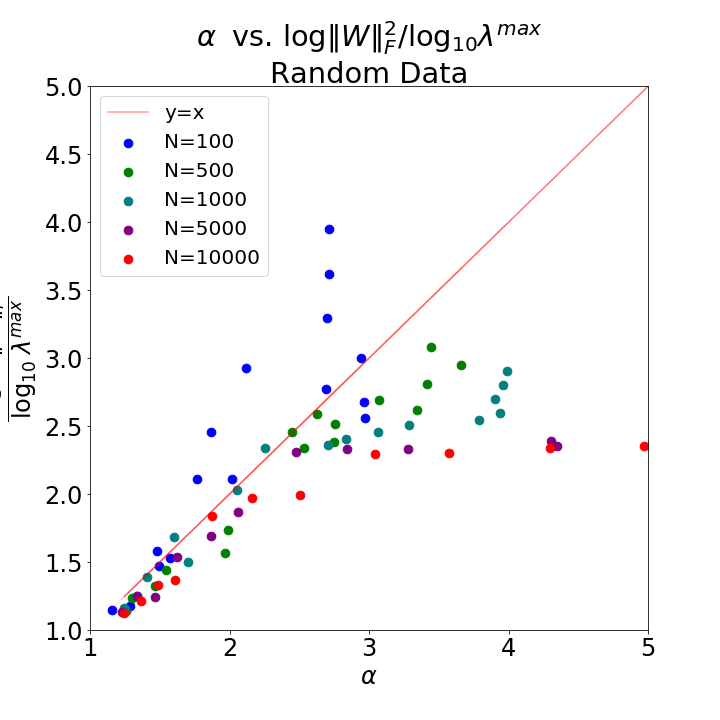}
   \caption{
            Numerical test of Eqn.~(\ref{eqn:basic_relation}) for random HT matrices across different HT Universality~classes.
           }
   \label{fig:randW}
\end{figure}

%% \michael{Make sure what I include in the above par is enough for what I need here.}
%% \michael{Here is the place to make explicit the connection between $\alpha$ and $\mu$, for finite $N$ and asymptotically, and to say Eqns.~(\ref{eqn:alpha_mu_vht_and_mht}) holds for both VHT and MHT, for realistic values of $N$.}

%To understand this relation better, and to sketch a proof,
%we will generate the data for a number of a Heavy-Tailed random matrices, 
%with different power law exponents $\mu$.
%Note that this linear relation holds over several log scales.  However,
%the relation does deviate from linearity at the smaller values of $\alpha\;\log_{10}\;\lambda^{max}$.
%This is readily explained below .  

%%MM%% This approximate relation formally only hold in the asymptotic limit of very small power law exponents $\alpha\rightarrow 1$ for
%%MM%% random heavy tailed matrices, but using Universality, we can safely extend it up to the finite-size MHT class, with
%%MM%% exponents $\alpha=4$ (and larger).  

%\paragraph{The PL--Norm Relation: Random Matrices versus Real Data.}
\section{The PL--Norm Relation: Random Matrices versus Real Data}
\label{sxn:appendix-random-vs-real}

\begin{figure}[!htb]
    \centering
    \subfigure[Random Pareto Matrices] {
        \includegraphics[scale=0.15]{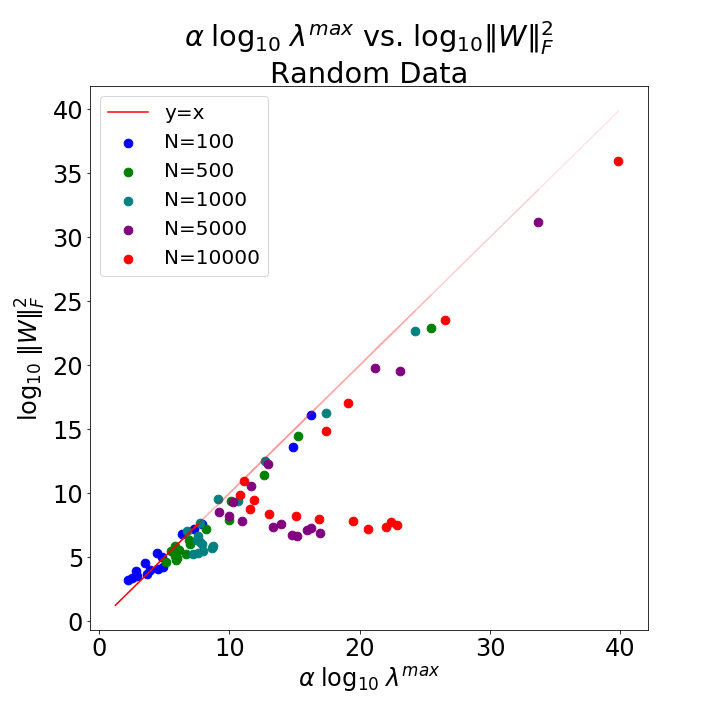} 
        \label{fig:relation-rand}
    }
    \subfigure[VGG11 Weight matrices]{
        \includegraphics[scale=0.15]{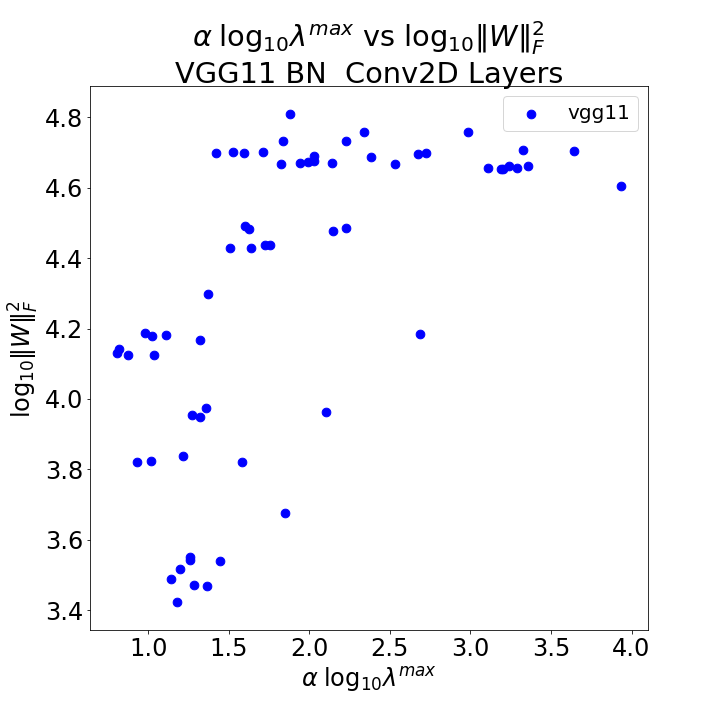} 
        \label{fig:relation-vgg11}
    }
        \caption{Relation between $\alpha\log_{10}\lambda^{max}$ and $\log_{10}\Vert\mathbf{W}\Vert^{2}_{F}$ for random (Pareto) matrices and real (VGG11) DNN weight matrices.
                 }
    \label{fig:relations}
\end{figure}

Here, we show, numerically, that Eqn.~(\ref{eqn:basic_relation}) holds qualitatively well and 
 more generally than the special case derived previously, including into the MHT Universality class, for both random and real data.
To illustrate this, we generate a large number of HT random matrices $\mathbf{W}^{rand}(\mu)$, with varying sizes $N$ (and $Q=1$), drawn from a Pareto distribution of Eqn.~(\ref{eqn:ht_dstbn}), with exponents $\mu\in[0.5, 5]$.
We then fit the ESD of each $\mathbf{W}^{rand}(\mu)$ to a PL using the method of Clauset et al.~\cite{CSN09_powerlaw,ABP14} to obtain the empirical exponent $\alpha$.   
Figure~\ref{fig:relation-rand} shows that there is a near-perfect relation between $ \alpha\log\lambda^{max}$ and $\log\Vert\mathbf{W}\Vert^{2}_{F} $, for this random data.
We also performed a similar PL fit for VGG11 weight matrices.
(See Section~\ref{sxn:emp} for some details on the VGG11 model.)
Figure~\ref{fig:relation-vgg11} shows the results, demonstrating for the VGG11 data an increasing relation until $ \alpha\log\lambda^{max} \approx 2.5$, and a saturation after that point.
Figure~\ref{fig:relations} illustrates
(among other things\footnote{Clearly, there are also differences between the HT random and the real DNN matrices, most notably that $ \alpha\log\lambda^{max} $ achieves much larger values for the random matrices.  This is discussed in more detail in Appendix~\ref{sxn:appendix-universality}.}):
that multiplying $\alpha$ by $\log\lambda^{max}$ leads to a relation that increases linearly with the (log of the squared) Frobenius norm for HT random matrices; that the two quantities are linearly correlated for real DNN weight matrices; and that both random HT and real, strongly-correlated matrices show similar saturation effects at large PL exponents.

%% \michael{Maybe comment also about how points large on the X axis have smaller $\alpha$.}
%% 
%% \charlesX{LEAD INTO HOW WE GOT THIS SECTION... PRESENT THE PL-NORM RELATION, SHOW THAT IT IS THE RIGHT SLOPE, AND HOW THE FINITE SIZE  EFFECTS LET US EXTEND THIS RELATION ACROSS UNIVERSALITY CLASSES IN ROUGH WAY, GIVING A USEFUL METRIC FOR ENGINEERING WORK.  REAL DATA IS SHOWN.  HAS 4 PLOTS.  USES NUMERICAL METHODS DESCRIBED ABOVE.  WE MAY WANT PSEUDOCODE ALSO ?}

\section{Random Pareto versus Non-random DNN Matrices} 
%\section{Heavy-Tailed Universality: Random Pareto versus Non-random DNN Matrices} 
\label{sxn:appendix-universality}

When we use Universality, as we do in our derivation of 
the basic PL--Norm Relation, 
%Eqn.~(\ref{eqn:basic_relation}),
we would like a method that applies both to HT random matrices as well as to non-random, indeed strongly-correlated, pre-trained DNN layer weight matrices that (as evidenced by their ESD properties) are in a HT Universality class.  
To accomplish this, however, requires some care: while the pre-trained $\mathbf{W}$ matrices do have ESDs that display empirical signatures of HT Universality~\cite{MM18_TR,MM19_HTSR_ICML}, they are \emph{not} random Pareto matrices.
Many of their properties, including their empirical Frobenius norms, behave very differently than that of a random Pareto matrix.  
(We saw this in Figure~\ref{fig:relations}, which showed that $ \alpha\log\lambda^{max} $ achieves much larger values for HT random matrices than real DNN weight matrices.)

To illustrate this, we generate a large number of HT random matrices $\mathbf{W}^{rand}(\mu)$, with exponents $\mu\in[0.5, 5]$, as described in Section~\ref{sxn:theory-new}.
We then fit the ESD of each $\mathbf{W}^{rand}(\mu)$ to a PL using the method of Clauset et al.~\cite{CSN09_powerlaw,ABP14} to obtain the empirical PL exponent $\alpha$. 
Figure~\ref{fig:fro-rand} displays the relationship between the (log of the squared) Frobenius norm and the $\mu$ exponents for these randomly-generated Pareto matrices.
(Similar but noisier plots would arise if we plotted this as a function of $\alpha$, due to imperfections in the PL fit.)
We did the same for the weight matrices (extracted from the Conv2D Feature Maps) from the pre-trained VGG11 DNN, again as described in Section~\ref{sxn:theory-new}.
Figure~\ref{fig:fro-vgg11} displays these results, here as a function of $\alpha$.

\begin{figure}[!htb]
   \centering
   \subfigure[Random Pareto Matrices] {
      \includegraphics[scale=0.15]{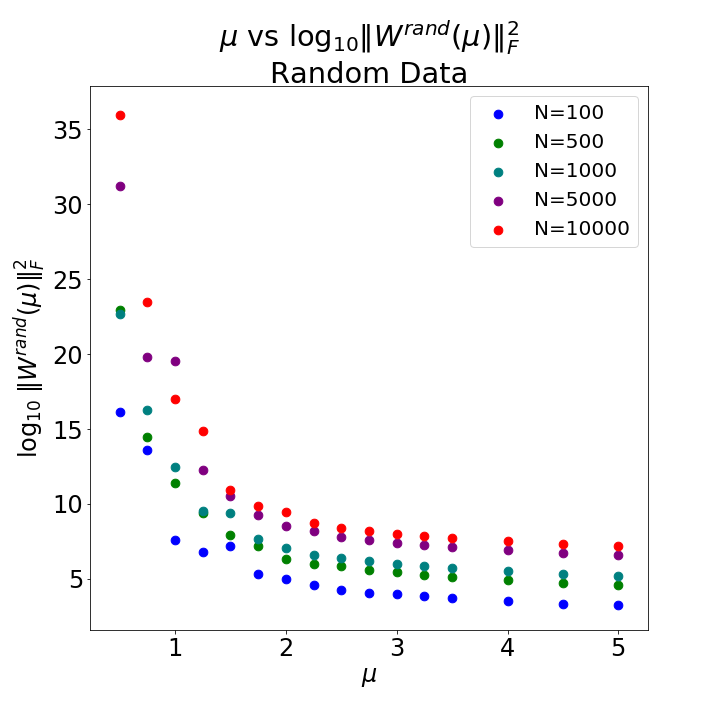} 
      \label{fig:fro-rand}
   }
   \subfigure[Pre-trained VGG11 Weight Matrices]{
      \includegraphics[scale=0.15]{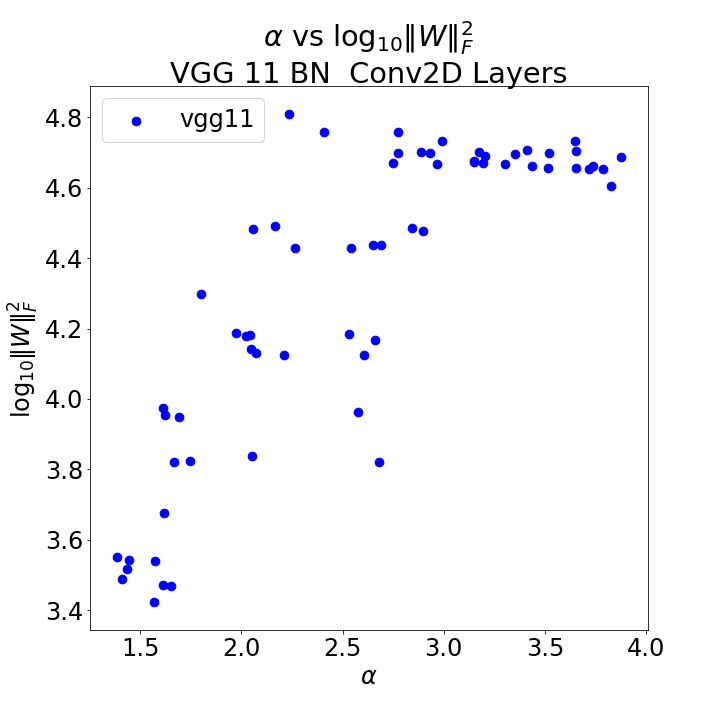} 
      \label{fig:fro-vgg11}
   }
   \caption{Dependence of Frobenius norm on PL exponents for random Pareto versus pre-trained DNN matrices.  }
   \label{fig:fnorm}
\end{figure}

From Figures~\ref{fig:fro-rand} and~\ref{fig:fro-vgg11}, we see that the properties of $\Vert\mathbf{W}\Vert^{2}_{F}$ differ strikingly for the random Pareto versus real/ron-random DNN weight matrices, and thus care must be taken when applying these Universality principles to strongly correlated systems.
For a random Pareto matrix, $\mathbf{W}^{rand}(\mu)$, the Frobenius norm $\Vert\mathbf{W}^{rand}(\mu)\Vert^{2}_{F}$ \emph{decreases with increasing exponent} $(\mu)$; and there is a modest finite-size effect.
(In addition, as the tails of the ESD $\rho(\lambda)$ get heavier, the largest eigenvalue $\lambda^{max}$ of $\mathbf{X}$ scales with the largest element of $\mathbf{W}^{rand}(\mu)$.) 
For the weight matrices of a pre-trained DNN, however, the Frobenius norm $\Vert\mathbf{W}\Vert^{2}_{F}$ \emph{increases with increasing exponent} $(\alpha)$, saturating at $\alpha\approx 3$.
This happens because, due to the training process, the $\mathbf{W}$ matrices themselves are highly-correlated, and not random matrices with a single large, atypical element.%
\footnote{This is easily seen by simply randomizing the elements of a real DNN weight matrix, and computing the ESD~again.}
In spite of this, the ESD $\rho(\lambda)$ of these pre-trained correlations matrices $\mathbf{X}$ display Universal HT behavior~\cite{MM18_TR,MM19_HTSR_ICML}.
In addition, as shown in Figure~\ref{fig:relation-vgg11}, Eqn.~(\ref{eqn:basic_relation}) is approximately satisfied, in the sense that $\alpha\log\lambda^{max}$ is positively correlated with $\log\Vert\mathbf{W}\Vert^{2}_{F} $.
This is one of the remarkable properties of HT-MU.

\section{Additional Discussion}
\label{sxn:appendix-addl-discussion}

We have presented an \emph{unsupervised} capacity control metric which predicts trends in test accuracies of a trained DNN---without peeking at the test data. 
This complexity metic, $\hat{\alpha}$ of Eqn.~(\ref{eqn:alpha_hat_specific}), is a weighted average of the PL exponents $\alpha$ for each layer weight matrix, where $\alpha$ is defined in the recent HT-SR Theory~\cite{MM18_TR,MM19_HTSR_ICML}, and where the weights are the largest eigenvalue $\lambda^{max}$ of the correlation matrix $\mathbf{X}$.  
We examine several commonly-available, pre-trained, production-quality DNNs by plotting $\hat{\alpha}$ versus the reported test accuracies.
This covers classes of DNN architectures including the VGG models, ResNet, DenseNet, etc. 
In nearly every class, and except for a few counterexamples, smaller $\hat{\alpha}$ corresponds to better average test accuracies, thereby providing a strong predictor of model quality.
We also show that this new complexity metric $\hat{\alpha}$ is approximately the average log of the squared Frobenius norm of the layer weight matrices, $\langle\log\Vert\mathbf{W}\Vert_{F}^{2}\rangle$, when accounting for finite-size effects:
$$
 \alpha\log\lambda^{max}\approx\log\Vert\mathbf{W}\Vert^{2}_{F}  .
$$
This provides an interesting connection between the Statistical Physics approach to learning (from Martin and Mahoney~\cite{MM17_TR,MM18_TR,MM19_HTSR_ICML}, that we extend here) and methods such as that of Liao et al.~\cite{LMBx18_TR}, who use norm-based capacity control metrics to bound worst-case generalization~error.

%% We should mention two higher-level comments.
%% %
%% First, 
%% one of the main insights of our approach is to highlight the importance of what might seem to be a technical issue to ignore, but which in our experience is \emph{extremely} important: the scaling or normalization for weight matrices used in the DNN, and how this relates to the difference between finite-size versus asymptotic effects.
%% This scaling issue has been highlighted perhaps most recently by Bartlett et al.~\cite{BFT17_TR} and Liao et al.~\cite{LMBx18_TR}.
%% The latter were interested in showing that classical generalization bounds can be tight---when normalization is performed appropriately.
%% Our approach complements theirs; but, to our knowledge, our approach is the first to highlight the connection with finite-size effects.
%% %
%% Second, 

It is worth emphasizing that 
we are taking a 
very %somewhat 
non-standard approach (at least for the DNN and ML communities) to address our main question.
We did not train/retrain lots and lots of (typically rather small) models, analyzing training/test curves, trying to glean from them bits of insight that might then extrapolate to more realistic models.
Instead, we took advantage of the fact that there already exist many (typically rather large) publicly-available pre-trained models, and we analyzed the properties of these models.
That is, we viewed these publicly-available pre-trained models as artifacts of the world that achieve state-of-the-art performance in computer vision, NLP, and related applications; and we attempted to understand why.
To do so, we analyzed the empirical (spectral) properties of these models; 
%from this, we formed a hypothesis as to why they perform well; 
and we then extracted data-dependent metrics to predict their generalization performance on production-quality models.
Given well-known challenges associated with training, 
%and given our results here as well as other recent results~\cite{MM18_TR,MM19_HTSR_ICML},
we suggest that this methodology be applied more generally.

Finally, one interesting aspect of our approach is that we can apply these complexity metrics \emph{across related DNN architectures}. 
This is in contrast to the standard practice in ML.
The equivalent notion would be to compare margins across SVMs, applied to the same data, but with different kernels. 
One loose interpretation is that a set of related of DNN models (i.e., VGG11, VGG13, etc.) is analogous to a single, very complicated kernel, and that the hierarchy of architectures is analogous to the hierarchy of hypothesis spaces in more traditional VC theory.
%\charlesX{more here ?  like this ?}
%\michael{Let's discuss this, to see what we can squeeze, given what is now popular.}   
Making this idea precise is clearly of interest.

We expect our result will have applications in the fine-tuning of pre-trained DNNs used for transfer learning, as in NLP and related applications.
Moreover, because we do not need to peek at the test data, our approach may prevent information from leaking from the test set into the model, thereby helping to prevent overtraining and making fined-tuned DNNs more robust.
Finally, our work also leads to a much harder theoretical question: is it possible to characterize properties of realistic DNNs to determine whether a DNN is overtrained---without peeking at the test data?

{ \iffalse

\newpage
\section{Appendix: Derivation of two relations}
\label{sxn:appendix-derivation-two-relations}

Here we derive the two relations, Eqns.~(\ref{eqn:alpha_mu_vht_and_mht}) and~(\ref{eqn:scaling_of_lambda_max}).
\michael{Maybe do this later.}

\fi }

{}

{}

\end{document}